\documentclass[letterpaper]{article}
\usepackage[ margin= 0.993 in]{geometry}
\usepackage{graphicx}
\usepackage[english]{babel}
\usepackage{fullpage}

\usepackage{amsmath, amsthm, amssymb}
\usepackage{amsfonts,bm}
\usepackage{dsfont}

\usepackage[labelfont=bf]{caption}
\usepackage{algorithm}
\usepackage{algorithmic}
\usepackage{thmtools}
\usepackage{thm-restate}
\usepackage{subcaption}

\newcommand{\figleft}{{\em (Left)}}

\newcommand{\figright}{{\em (Right)}}

\def\1{\bm{1}}

\def\vtheta{{\bm{\theta}}}

\def\vh{{\bm{h}}}

\def\vx{{\bm{x}}}

\def\vz{{\bm{z}}}

\DeclareMathAlphabet{\mathsfit}{\encodingdefault}{\sfdefault}{m}{sl}
\SetMathAlphabet{\mathsfit}{bold}{\encodingdefault}{\sfdefault}{bx}{n}

\newcommand{\KL}{D_{\mathrm{KL}}}

\DeclareMathOperator*{\argmin}{arg\,min}

\newcommand{\norm}[1]{\left\|#1\right\|}
\newcommand{\abs}[1]{\lvert#1\rvert}

\DeclareMathOperator{\poly}{\mathrm{poly}}

\newcommand{\anchorterm}[2]{%
  \hypertarget{#1}{\text{#2}}
  \expandafter\gdef\csname anchortext@#1\endcsname{\text{#2}}%
}

\DeclareRobustCommand{\termlink}[1]{%
  \hyperlink{#1}{\text{\csname anchortext@#1\endcsname}}%
}

\newcommand{\ind}{\mathds{1}}
\newcommand{\op}{\mathrm{op}}

\makeatletter
    \newcommand{\anchortext@termtau}{\text{(I)}}
    \newcommand{\anchortext@challengeweight}{\text{Challenge 1}}
    \newcommand{\anchortext@termIIi}{\text{(II.i)}}
    \newcommand{\anchortext@termIIii}{\text{(II.ii)}}
\makeatother

\newcommand{\Dmax}{D_{\max}}

\usepackage{enumitem}
\newlist{tightenum}{enumerate}{1}
\setlist[tightenum]{
  label=\arabic*.,
  leftmargin=13pt,
  labelindent=0pt,
  listparindent=0pt,
  itemsep=-3pt,
  topsep=-15pt,
}

\newcommand{\bx}{\vx}
\newcommand{\bz}{\vz}
\newcommand{\btheta}{\vtheta}
\newcommand{\bpi}{\vpi}
\newcommand{\bh}{\vh}

\newcommand{\vphi}{\varphi}
\newcommand{\Dth}{D_{\theta}}
\newcommand{\Bth}{B_{\theta}}
\newcommand{\Lam}{\Lambda}
\newcommand{\Ag}{\mathsf{A}_{\mathrm{g}}}

\newcommand{\E}{\ensuremath{\mathbf{E}}}

\newcommand{\R}{\ensuremath{\mathbb{R}}}
\renewcommand{\L}{\ensuremath{\mathcal{L}}}

\newcommand{\bmu}{\ensuremath{\bm{\mu}}}

\newcommand{\ip}[1]{\ensuremath{\left\langle #1 \right\rangle}}

\newcommand{\pimin}{\ensuremath{\pi_{\min}}}

\newcommand{\vpi}{\ensuremath{\bm{\pi}}}

\newcommand{\cL}{\ensuremath{\mathcal{L}}}
\newcommand{\cP}{\ensuremath{\mathcal{P}}}

\usepackage{thmtools,thm-restate,apptools}

\theoremstyle{plain}
\newtheorem{theorem}{Theorem}[section]
\newtheorem{lemma}{Lemma}[section]

\newtheorem{proposition}{Proposition}[section]
\newtheorem{claim}{Claim}[section]

\newtheorem{remark}{Remark}[section]
\newtheorem{assumption}{Assumption}[section]
\newtheorem*{conjecture*}{Conjecture}

\usepackage{xcolor}
\usepackage{float}
\usepackage{url}
\usepackage{tabularx}
\usepackage{array}
\usepackage{booktabs}
\usepackage{natbib}
\setcitestyle{authoryear,round,citesep={;},aysep={,},yysep={;}}
\usepackage[colorlinks=true,pdfborder={0 0 0}]{hyperref}

\makeatletter\newcommand{\Loj}{{\csname\cf@encoding\string\L\endcsname}ojasiewicz}\makeatother

\begin{document}
\title{Provably Learning Mixtures of \\ Linear Experts}
\author{Adrian Candocia, Aamod Varma, and Mark Davenport}
\date{}
\maketitle

\begin{abstract}
Mixture-of-Experts (MoE) models are widely deployed in modern machine learning. However, few convergence guarantees exist for gradient-based training of MoEs: the objective is nonconvex, and even simple mixtures trap first-order methods in spurious local optima. We analyze a canonical gated mixture featuring linear experts and a gate induced by Gaussian clusters in the input space, and give, to our knowledge, the first global convergence guarantee for jointly training the gate and the experts of such a model by gradient EM on its likelihood. Given a logarithmic overparameterization relative to the true number of experts $m$, gradient EM converges from randomly initialized gates and least-squares-initialized experts to any prescribed accuracy in polynomially many iterations, under a separation condition on the gate centers. Our analysis establishes a gradient-dominance inequality for the coupled dynamics by decoupling the loss into per-cluster subproblems. Empirical results validate the predicted $m\log m$ phase transition and suggest that the least-squares warm start required by our analysis is not needed in practice: randomly initialized experts perform nearly as well. We also find that leaving the mixing weights unconstrained concentrates routing onto the true number of experts, whereas load-balancing penalties spread it, so that the fitted model can be pruned to a fraction of its width without loss of accuracy.
\end{abstract}
{\let\thefootnote\relax\footnote{{A. Candocia, A. Varma, and M. Davenport are with the School of Electrical and Computer Engineering, Georgia Institute of Technology, Atlanta, GA 30332, USA. \\ E-mails: \texttt{acandocia3@gatech.edu, aamodvarma@gatech.edu, mdav@gatech.edu}}}}

\section{Introduction}
\label{sec:intro}

Mixture-of-experts (MoE) models decompose a prediction problem across a collection of specialized \emph{experts}, with a \emph{gate} (or router) that decides which experts are responsible for each input \citep{jacobs1991,jordan1994}. The idea is now standard in large models: sparsely gated MoE layers let the parameter count of a network grow by orders of magnitude while the compute per input stays fixed \citep{shazeer2017,fedus2022}. Often, many more experts than are necessary are initialized, the gate is initialized at random, and expert load-balancing penalties are implemented to incentivize broad expert utilization. Each of these choices is motivated by experience rather than theory. Why gradient-based training of a gated mixture should succeed at all is only partly understood.

An MoE can be seen as a latent-variable model whose likelihood is nonconvex. The nonconvexity comes from the coupling of gate and experts: the gate decides which expert sees which data, and the experts' errors feed back into the gate. The theory of expectation--maximization (EM) has shed considerable light on the specific latent-variable model of a Gaussian mixture model (GMM). When the GMM has exactly as many components as the ground-truth, EM and its gradient variants enjoy global guarantees only for two components \citep{xu2016global,daskalakis2017} or from a good initialization \citep{balakrishnan2017,yan2017}, and for three or more Gaussian components the population likelihood provably has spurious local maxima that trap EM from random initialization with high probability \citep{jin2016}. A recent line of work has shown that this obstruction disappears when the model is \emph{over-parameterized}: with $n=\Omega(m\log m)$ fitted components, randomly initialized gradient EM recovers an $m$-component Gaussian mixture in polynomial time, first for $m=1$ \citep{xu2024} and then for general well-separated mixtures \citep{zhou2026}. Recent work on MoEs over-provision  of experts so that every latent cluster is covered at initialization is the mechanism behind recent gradient-based analyses \citep{chen2022moe,kawata2025,liao2026}. Whether it carries the joint likelihood dynamics of gate and experts has remained open.

In this paper we give a global convergence guarantee for gradient-based training of a Gaussian-gated mixture of linear experts. Data are generated by drawing a latent expert index $\ell$ with probability $\pi^*_\ell$, then an input $\bx\sim N(\bmu^*_\ell,I_d)$, and finally a response $y=\bx^\top\btheta^*_\ell+\epsilon$ with $\epsilon\sim N(0,1)$ independent of $(\ell,\bx)$. Although this is a generative model, its predictive function is exactly an MoE in the sense used in practice. A quick calculation shows that
\begin{equation}\label{eq:moe-intro}
\E[y\mid\bx]=\sum_{i=1}^n \mathrm{softmax}_i\big(\ip{\bx,\bmu_i}+b_i\big)\,\bx^\top\btheta_i,\qquad b_i=\log\pi_i-\tfrac12\|\bmu_i\|^2,
\end{equation}
that is, a softmax router with weights $\bmu_i$ and biases $b_i$ with linear experts $\btheta_i$. The mixing weights $\pi_i$, which are usually fixed or ignored in MoE analysis, play the role of learnable router biases. We train this model by gradient EM on the population log-likelihood, which we show coincides with gradient descent on the Kullback--Leibler (KL) divergence between the truth and the model.

Our main result (Theorem~\ref{thm:global}) states that when the fitted model has $n\gtrsim m\log m$ experts, the gates are initialized at random from the data, the experts are initialized by a windowed least-squares fit, and the true gate centers are well separated, gradient EM drives the population KL loss below any target $\varepsilon\ge e^{-c\,dn}$ within a polynomial number of steps. Global guarantees for gradient-based MoE training exist for staged algorithms, purpose-built losses and specific teacher models (Section~\ref{sec:related}); to our knowledge this is the first for joint training of gate and experts on the likelihood itself, and the first in which the mixing weights are free. The proof establishes a gradient-dominance inequality. A central difficulty is controlling the expert parameters: the expert errors (residuals) $y-\bx^\top\btheta_i$ have to remain bounded within a shrinking budget throughout training. The least-squares warm start supplies the initial bound on the residuals, and an expert-drift argument ensures the budget is not violated.

Our contributions are as follows.
\begin{itemize}[nosep,leftmargin=*]
\item \textbf{A global guarantee for joint gate--expert training} (Theorem~\ref{thm:global}): with $n\gtrsim m\log m$ experts (for balanced mixing weights), random gates and least-squares experts, population gradient EM reaches any accuracy $\varepsilon\ge e^{-c\,dn}$ in polynomially many steps.
\item \textbf{A proof technique} that decouples the loss into per-cluster problems, controls each by a Gaussian log-Sobolev inequality, and keeps the expert residuals inside a shrinking budget along the whole trajectory.
\item \textbf{Experiments} showing (i) a sharp success transition along $n\asymp m\log m$ (Figure~\ref{fig:heatmap}), (ii) that randomly initialized experts succeed almost as often as least-squares experts once the gates cover the clusters (Figure~\ref{fig:warm-start-ablation}), and (iii) that free mixing weights concentrate routing on about $m$ experts and make the model prunable, while load-balancing penalties destroy both properties (Figures~\ref{fig:balance-eff} and~\ref{fig:pruning-kl}).
\end{itemize}

\section{Related Work}
\label{sec:related}

\paragraph{EM for exactly parameterized mixtures.}
The convergence of EM for Gaussian mixtures was studied early by \citet{Xu1996}. Modern convergence theory begins with \citet{balakrishnan2017}, who showed that population EM converges locally at a geometric rate, and verified this for two-component Gaussian mixtures and two-component mixed linear regression. Subsequent work extended these guarantees to mixtures of many linear regressions, to two-component mixed linear regression from random initialization, and to well-separated Gaussian mixtures \citep{kwonMLR2020,kwonMLR2024,kwon2020gmm}. For three or more components the picture changes: \citet{jin2016} showed that the population log-likelihood has spurious local maxima to which EM from random initialization converges with high probability.

\paragraph{Over-parameterized mixtures.}
Fitting more components than the data contain changes the picture. \citet{dwivedi2018} showed that over-specification creates singularities in the Fisher information and slows the local rate of EM from geometric to polynomial, and \citet{xu2024} then proved the first global result in this regime: gradient EM fitting an $n$-component mixture to a single Gaussian converges from random initialization at rate $O(1/\sqrt t)$. \citet{zhou2026} extended this to a general $m$-component, well-separated Gaussian mixture, showing that $n=\Omega(m\log m)$ components suffice for randomly initialized gradient EM to converge globally in polynomial time and with polynomially many samples. Our analysis builds on their framework; the experts add a second parameter set whose errors enter the gate through the regression residuals, and controlling that coupling is the main new difficulty.

\paragraph{Theory of mixtures of experts.}
Mixtures of experts were introduced by \citet{jacobs1991} and \citet{jordan1994}, who already proposed EM for training them. A recent body of work analyzes the \emph{statistical} properties of MoE estimators: convergence rates of the maximum-likelihood estimator for Gaussian MoEs \citep{ho2022moe} along with the effect of the softmax gate on identifiability and rates \citep{nguyen2023demystifying,nguyen2024general}. The Gaussian-gated model of \citet{nguyen2024gaussiangated} is closest to ours; however, these results concern the estimator rather than the algorithm that computes it. On the algorithmic side, \citet{fruytier2025} cast EM for mixtures of experts as mirror descent and obtain local guarantees. \citet{makkuva2019} recover the experts by a moment method and then the gate by EM, and \citet{makkuva2020} obtain global guarantees for gradient descent by designing separate losses for the gate and the experts. Closest to us are analyses of gradient-based training under a latent cluster structure: \citet{chen2022moe} (top-1 routing and classification, where the nonlinearity of the experts is essential), \citet{kawata2025} (Gaussian clusters with single-index targets, trained by a multi-phase SGD schedule that updates the router and the experts separately), and \citet{liao2026} (joint gradient flow on the population squared error for a normalized sigmoid router with Hermite experts, an orthonormal teacher and $d\ge\mathrm{poly}(m)$, followed by pruning). All three provision $\Omega(m\log m)$ experts so that every cluster is covered at initialization, as we do. We differ in analyzing the likelihood objective directly, with gate, experts and mixing weights trained jointly by one gradient method, for arbitrary well-separated centers and mixing weights and without a high-dimensional regime.

\paragraph{Routing and load balancing in practice.}
Sparsely gated MoE layers \citep{shazeer2017} and their simplifications \citep{fedus2022} route each input to a few experts and add an auxiliary loss that penalizes imbalanced expert utilization, since imbalanced routing wastes capacity and complicates parallel execution. In the idealized case where the fraction of inputs dispatched to expert $i$ equals its average routing probability $\bar\rho_i$, the Switch auxiliary loss \citep{fedus2022} equals $n\sum_{i=1}^n\bar\rho_i^2=n/n_{\mathrm{eff}}$, where $n_{\mathrm{eff}}:=\big(\sum_{i=1}^n\bar\rho_i^2\big)^{-1}$ is the effective number of experts the router uses; minimizing the penalty therefore pushes utilization toward the full fitted width $n$. More recently, balancing has also been pursued through per-expert bias terms that are added to the routing scores used for expert selection and adjusted by a rule during training, instead of through an auxiliary loss \citep{wang2024,deepseekv3}. These biases still aim at balanced load, but they leave the gradient objective untouched.

\section{Learning a Mixture of Experts}
\label{sec:model}

\subsection{Problem Statement}
Our data model is an $m$-component mixed linear regression in which each regression vector has its own Gaussian covariate cluster. A sample $\bz=(\bx,y)\in\R^d\times\R$ is generated by
\[
\ell\sim\text{Cat}(\pi_1^*,\dots, \pi_m^*),\qquad \bx\mid\ell\sim N(\bmu^*_\ell,I_d),\qquad y\mid\bx,\ell\sim N(\bx^\top\btheta^*_\ell,1).
\]
We can now define the mixture distribution $p_*$ over $\bz$. Let $\vphi(\bx;\bm{\alpha}):=(2\pi)^{-d/2}e^{-\|\bx-\bm{\alpha}\|^2/2}$ denote the p.d.f. of a Gaussian with mean $\bm\alpha\in\R^d$ and unit covariance and write the one dimensional case as $\vphi_1(y;a)$. Also, let $\cP^{k}\subset\R^{k+1}$ denote the probability simplex. For $\bmu\in\R^d$ and $\btheta\in\R^d$ the joint density of one component is $\vphi(\bz;\bmu,\btheta):=\vphi(\bx;\bmu)\,\vphi_1(y;\bx^\top\btheta)$.
The ground truth is then the $m$-component joint density
\begin{equation}\label{eq:truth}
p_*(\bz)=\sum_{\ell=1}^m\pi^*_\ell\,\vphi(\bz;\bmu^*_\ell,\btheta^*_\ell), \quad \bpi^*\in\cP^{m-1}
\end{equation}
and the over-parameterized model has $n\ge m$ experts,
\begin{equation}\label{eq:model}
p(\bz)=p(\bz;\bpi,\bmu,\btheta)=\sum_{i=1}^n\pi_i\,\vphi(\bz;\bmu_i,\btheta_i),\qquad \bpi\in\cP^{n-1}.
\end{equation}
We write $\vphi_i(\bz):=\vphi(\bz;\bmu_i,\btheta_i)$ and $\vphi^*_\ell(\bz):=\vphi(\bz;\bmu^*_\ell,\btheta^*_\ell)$, and drop parameter arguments when they are clear from context.
Define $\pimin:=\min_\ell\pi^*_\ell$. We write $\E_\ell:=\E_{\bz\sim\vphi^*_\ell}$ for expectation under the $\ell$th true component, i.e., $\bx\mid\ell\sim N(\bmu^*_\ell,I_d)$ and then $y\mid\bx,\ell\sim N(\bx^\top\btheta^*_\ell,1)$, so that $\E_{p_*}=\sum_\ell\pi^*_\ell\E_\ell$.

Let $S_\ell:=\{i\in[n]:\|\bmu_i-\bmu^*_\ell\|\le\|\bmu_i-\bmu^*_j\|\ \forall j\}$ (ties broken by a fixed rule) be the Voronoi group of $\bmu^*_\ell$, $\ell(i)$ the group of $i$, $\hat\pi_\ell:=\sum_{i\in S_\ell}\pi_i$ and
\begin{equation}\label{eq:potentials}
U_\mu(\bmu):=\sum_{\ell}\sum_{i\in S_\ell}\|\bmu_i-\bmu^*_\ell\|^2,\qquad
U_\theta(\bmu,\btheta):=\sum_{\ell}\sum_{i\in S_\ell}\|\btheta_i-\btheta^*_\ell\|^2,\qquad U:=U_\mu+U_\theta
\end{equation}
measure how far the fitted components sit from the truth of their group.

\subsection{EM Algorithm}
EM iteratively updates the parameters $\bmu:=(\bmu_1^\top,\dots,\bmu_n^\top)^\top$, $\btheta:=(\btheta_1^\top,\dots,\btheta_n^\top)^\top$ and $\bpi:=(\pi_1,\dots,\pi_n)$ of $p$. At the current iterate $(\bpi',\bmu',\btheta')$, the \textbf{E-step} computes the posterior probability that component $i$ generated $\bz$,
\[
\psi_i(\bz;\bpi',\bmu',\btheta'):=\frac{\pi'_i\,\vphi(\bz;\bmu'_i,\btheta'_i)}{p(\bz;\bpi',\bmu',\btheta')},
\]
written $\psi_i(\bz)$ when the parameters are clear, and the \textbf{M-step} maximizes, with these posteriors frozen, the expected complete-data log-likelihood (up to an additive constant)
\[
Q(\bpi,\bmu,\btheta\mid\bpi',\bmu',\btheta'):=\E_{p_*}\Big[\sum_{i=1}^n\psi_i(\bz;\bpi',\bmu',\btheta')\big(\log\pi_i-\tfrac12\|\bx-\bmu_i\|^2-\tfrac12(y-\bx^\top\btheta_i)^2\big)\Big].
\]
The expectation over $p_*$ makes this \emph{population} EM. We analyze the \emph{gradient EM} variant in which the M-step for $(\bmu,\btheta)$ is replaced by one gradient step on $Q$, while the mixing weights are updated by approximately minimizing the loss $\cL$ below over the simplex, a convex problem (Algorithm~\ref{alg:main}). By Claim~\ref{claim:gradEM}, the gradient step on $Q$ is a gradient step on $\cL$.

\subsection{Loss}
We use the KL divergence loss, as in \citet{xu2024,zhou2026},
\begin{equation}\label{eq:loss}
\cL(\bpi,\bmu,\btheta):=\KL(p_*\|p)=\E_{p_*}[-\log p(\bz)]-\E_{p_*}[-\log p_*(\bz)]
\end{equation}
There are two reasons for this choice. First, $\cL$ is nonnegative and vanishes exactly when $p=p_*$, i.e., when the model reproduces the data density and hence the true regression function $\E_{p_*}[y\mid\bx]$. Second, gradient EM is gradient descent on $\cL$:
\begin{claim}\label{claim:gradEM}
For any fixed $(\bpi',\bmu',\btheta')$, $\nabla_{\bmu,\btheta}Q(\bpi,\bmu,\btheta\mid\bpi',\bmu',\btheta')\big|_{(\bpi',\bmu',\btheta')}=-\nabla_{\bmu,\btheta}\cL(\bpi',\bmu',\btheta')$.
\end{claim}
As is common when training and deploying MoE models, we are concerned with minimizing the loss $\cL$ via a gradient descent procedure. Exact parameter recovery
is not the focus of this work.

\section{Results}
\label{sec:results}

\subsection{Assumptions}
As is standard, our guarantees are stated in terms of the separation and spread of the true parameters:
\[
\Delta:=\min_{\ell\ne j}\|\bmu^*_\ell-\bmu^*_j\|, \quad \Dmax:=\max_{\ell\ne j}\|\bmu^*_\ell-\bmu^*_j\|, \quad \Dth:=\max_{\ell\ne j}\|\btheta^*_\ell-\btheta^*_j\|
\]
The loss and the updates are translation equivariant, so we may assume $\max_{\ell}\|\bmu^*_\ell\|\le\Dmax$ (translate so that one true center is the origin). With $\Lam:=d+\Dmax^2$ this gives $\E_\ell\|\bx\|^2\le2\Lam$ for each $\ell$.
We then make the following well-separatedness assumption on \textit{only} the minimum distance between the centers, where $\delta\in(0,1)$ is the failure probability of Lemma~\ref{lem:init} and Theorem~\ref{thm:global}.
\begin{assumption}[Well-separated gates]\label{ass:sep}
$m\ge2$ and
\[
\Delta\ge C_{\mathrm{sep}}\max\Big\{\sqrt{\log(dnm(1+\Dmax)(1+\Dth)/\pimin)},\ \sqrt{n(d+\log(n/\delta))}\Big\}
\]
for a large universal constant $C_{\mathrm{sep}}$. (When $m=1$ all cross-cluster terms below vanish and the separation requirements are vacuous.)
\end{assumption}
The term $\sqrt{n(d+\log(n/\delta))}\asymp B$ is the radius of the region $U\le B^2/2$ that the iteration never leaves, and the proof needs every fitted gate to stay far from every foreign true center, i.e., $\Delta\gg B$.
We highlight that a sharper separation condition defined in the appendix, \eqref{eq:sepD1}, is used in the proofs. Throughout the text, $c,C>0$ denote universal constants whose values may change from line to line, $a\lesssim b$ means $a\le Cb$, and
we set
\[
\rho_0:=C\sqrt{d+\log(4n/\delta)},\qquad B^2:=4n\rho_0^2,\qquad L_0:=\log n+2(d+1),
\]
and, for a target accuracy $\varepsilon$, $\Bth:=c\min\{\varepsilon/B,\,1/((1+\Dth)\Lam)\}$. Here $\rho_0$ bounds the distance of every initial gate from the true center it was drawn from, $B$ is the radius of the region in which the analysis takes place, $L_0$ bounds the initial loss, and $\Bth$ is the expert-error budget. Finally, $\Ag,A=\mathrm{poly}\big(d,m,n,\tfrac1{\pimin},B,1+\Dmax,1+\Dth\big)$ denote polynomial prefactors whose degree is universal and which may be enlarged from line to line (exact forms in Appendix~\ref{sec:appconst}).

\subsection{Initialization \& Algorithm}
Our analysis needs the experts to start close to the truth. We obtain this by a windowed least-squares fit: each $\btheta^{(0)}_i$ regresses $y$ on $\bx$ using only data within distance $R$ of $\bmu^{(0)}_i$,
\begin{equation}\label{eq:lsinit}
\btheta^{(0)}_i:=\argmin_{\btheta\in\R^d}\ \E_{p_*}\Big[\ind\{\|\bx-\bmu^{(0)}_i\|\le R\}\,(y-\bx^\top\btheta)^2\Big].
\end{equation}
When $\Delta$ is large, the window contains almost only data from the cluster that $\bmu^{(0)}_i$ was drawn from, so \eqref{eq:lsinit} nearly returns that cluster's regression vector; this is item~3 of the following lemma:

\begin{lemma}[Initialization (Informal)]\label{lem:init}
Let Assumption~\ref{ass:sep} hold, fix $\delta\in(0,1)$, suppose $n\ge\frac{C}{\pimin}\log\frac{2m}{\delta}$, and let $R:=\rho_0+2\sqrt d$ (so that $\Delta\ge4(R+\rho_0)$ under Assumption~\ref{ass:sep}). Write $J_i$ for the cluster from which $\bmu_i^{(0)}$ was drawn in Algorithm~\ref{alg:main}. Then with probability at least $1-\delta$:
\begin{enumerate}[nosep]
\item for each $\ell\in[m]$ there is $i$ with $J_i=\ell$ and $\|\bmu^{(0)}_i-\bmu^*_\ell\|\lesssim \sqrt d$;
\item for every $i$, $\|\bmu^{(0)}_i-\bmu^*_{J_i}\|\le\rho_0$;
\item for every $i$, $\|\btheta^{(0)}_i-\btheta^*_{J_i}\|\le\Ag e^{-c\Delta^2}$;
\item the initial loss satisfies $\cL^{(0)}\le L_0$ (using item~3).
\end{enumerate}
\end{lemma}
Item~1 guarantees that every true cluster receives a gate, item~2 bounds the distance of every gate to the center it was drawn from, item~3 is the accuracy of the warm start, and item~4 bounds the initial loss. For $\pimin\gtrsim1/m$ (in particular for balanced weights) the requirement on $n$ reads $n\gtrsim m\log(m/\delta)$: this is where the $m\log m$ over-parameterization comes from, and it is a coupon-collector bound, since each of the $m$ true clusters must be hit by at least one of $n$ random gate draws.

Algorithm~\ref{alg:main} is gradient EM with extra care for the mixing weights $\bpi$, which are constrained to the simplex $\cP^{n-1}$. The weight step is $\varepsilon_\pi$-optimal, i.e., it has simplex dual gap at most $\varepsilon_\pi$ and does not increase the loss (Appendix~\ref{sec:dualgap}); the EM fixed-point iteration $\pi_i\leftarrow\E_{p_*}\psi_i$ run to that tolerance qualifies. Near-optimality is what controls the group masses $\hat\pi_\ell$ (Lemma~\ref{lem:sk-lossKL}); we do not analyze a projected gradient step on $\bpi$.

\begin{algorithm}[t]
\caption{Population gradient EM for the MoE with least-squares warm start}\label{alg:main}
\begin{algorithmic}[1]
\STATE \textbf{Input:} fitted width $n$, step size $\eta$, horizon $T$, weight-subproblem tolerance $\varepsilon_\pi$, window radius $R$
\STATE \textbf{(Random) gate initialization:} draw $\bmu^{(0)}_1,\dots,\bmu^{(0)}_n\overset{\text{i.i.d.}}{\sim}p_*^{\bx}$ (the $\bx$-marginal of $p_*$) and set $\pi^{(0)}_i=1/n$
\STATE \textbf{Expert initialization (windowed least squares):} set each $\btheta^{(0)}_i$ as in \eqref{eq:lsinit}
\FOR{$t=0$ to $T-1$}
    \STATE $\bpi^{(t+1)}\gets$ an $\varepsilon_\pi$-optimal, loss non-increasing solution of the convex problem $\min_{\bpi\in\cP^{n-1}}\cL(\bpi,\bmu^{(t)},\btheta^{(t)})$
    \STATE $(\bmu^{(t+1)},\btheta^{(t+1)})\gets(\bmu^{(t)},\btheta^{(t)})-\eta\,\nabla_{\bmu,\btheta}\cL(\bpi^{(t+1)},\bmu^{(t)},\btheta^{(t)})$
\ENDFOR
\STATE \textbf{Output:} $(\bpi^{(T)},\bmu^{(T)},\btheta^{(T)})$
\end{algorithmic}
\end{algorithm}

\subsection{Main Result}

We now state an informal version of the main result and sketch the argument; full proofs are in the appendix.

\begin{theorem}[Convergence of gradient EM (Informal)]\label{thm:global}
Let Assumption~\ref{ass:sep} hold, fix $\delta\in(0,1)$, and fix a target accuracy $\varepsilon$ with
$e^{-c\,dn}\le\varepsilon\le c\,\pimin$. Run Algorithm~\ref{alg:main}
with $\eta\le c\min\{1/A,\ \Bth^2/L_0\}$ and $\varepsilon_\pi\le c\,\varepsilon/\Ag$. Then with probability at least
$1-\delta$ over the random initialization,
\[
\cL(\bpi^{(t)},\bmu^{(t)},\btheta^{(t)})\le\varepsilon
\quad\text{for some } t\le T:=\frac{CB^2}{\eta\varepsilon},
\]
and the loss is non-increasing up to that time. For the largest admissible step size, $T = O(L_0B^4/\varepsilon^3)$. Moreover, for every $t\le T$ before the target is reached, with $B$ and $\Bth$ as above,
\[
U(\bmu^{(t)},\btheta^{(t)})\le\frac{B^2}2,
\qquad
U_\theta(\bmu^{(t)},\btheta^{(t)})\le\Bth^2,
\qquad\text{and the groups } S_\ell \text{ do not change.}
\]
\end{theorem}

We prove Theorem~\ref{thm:global} by establishing a gradient-dominance inequality $\|\nabla_{\bmu,\btheta}\cL\|\gtrsim\cL/B$ on the region $U\le B^2/2$, $U_\theta\le\Bth^2$ (a \Loj\ inequality). We obtain it by lower bounding the projection of the gradient onto the \emph{intuitive} direction that moves every gate toward the true center of its group and every expert toward the true regression vector of that group:
\[
P:=\sum_{\ell\in[m]}\sum_{i\in S_\ell}
\Big(\ip{\nabla_{\bmu_i}\cL,\bmu_i-\bmu^*_\ell}+\ip{\nabla_{\btheta_i}\cL,\btheta_i-\btheta^*_\ell}\Big)
\]

When the separation is large, each cluster is essentially an isolated problem: $\|\bmu_i-\bmu^*_\ell\|\le B$ for $i\in S_\ell$, while $\|\bmu_i-\bmu^*_j\|\ge\Delta-B\gg B$ for $j\ne\ell$. Since Gaussian densities decay exponentially, a component of group $\ell$ is essentially invisible to data from cluster $j$:

\begin{lemma}[Cross-group terms]\label{lem:sk-cross}
For $i\in S_\ell$, $j\ne\ell$ and any nonnegative polynomial weight $W(\bz)$ of bounded degree (Lemma~\ref{lem:ratio}),
$\E_{j}\big[\psi_i(\bz)W(\bz)\big]\le\Ag e^{-c\Delta^2}$.
\end{lemma}

We now proceed by writing the normalized within-group mixture, its posterior, and the posterior-averaged weights
\[
\lambda^{(\ell)}_i:=\frac{\pi_i}{\hat\pi_\ell},\qquad
q_\ell(\bz):=\sum_{i\in S_\ell}\lambda^{(\ell)}_i\vphi_i(\bz),\qquad
\psi^{(\ell)}_i:=\frac{\lambda^{(\ell)}_i\vphi_i}{q_\ell},\qquad
\bar\lambda^{(\ell)}_i:=\E_\ell\big[\psi^{(\ell)}_i\big],
\]
so that $\cL_\ell:=\KL(\vphi^*_\ell\|q_\ell)$ is the loss of the $\ell$th true component. Grouping the model's components by the $S_\ell$ writes both densities as $m$-term mixtures: $p_*=\sum_\ell\pi^*_\ell\vphi^*_\ell$ and $p=\sum_\ell\hat\pi_\ell q_\ell$. The log-sum inequality gives
$\KL(p_*\|p)\le\sum_\ell\pi^*_\ell\KL(\vphi^*_\ell\|q_\ell)+\KL(\bpi^*\|\hat\bpi)$. Separation and the near-optimal $\bpi$-step make the weight-mismatch term small:

\begin{lemma}[Loss vs.\ within-group KL]\label{lem:sk-lossKL}
$\cL\le\sum_{\ell}\pi^*_\ell\cL_\ell+C\varepsilon_\pi+\Ag e^{-c\Delta^2}$.
\end{lemma}

Turning our attention to the individual clusters, define the within-cluster projection
\begin{equation}\label{eq:Pell-main}
P_\ell:=\sum_{i\in S_\ell}\Big(\ip{\nabla_{\bmu_i}\cL_\ell,\bmu_i-\bmu^*_\ell}
+\ip{\nabla_{\btheta_i}\cL_\ell,\btheta_i-\btheta^*_\ell}\Big).
\end{equation}

Using Stein's lemma \citep{stein1981} (Lemma~\ref{lem:stein}) and the Gaussian log-Sobolev inequality \citep{gross1975}, we show that within a cluster the projection $P_\ell$ dominates the cluster loss $\cL_\ell$:

\begin{lemma}[Projection comparison]\label{lem:sk-entropy}
Under the assumptions of Theorem~\ref{thm:global}, for every $\ell\in[m]$,
\[
P_\ell\ \ge\ 2\cL_\ell-2\KL(\bar\lambda^{(\ell)}\|\lambda^{(\ell)})-c\,\varepsilon,
\qquad
\sum_\ell\pi^*_\ell\KL(\bar\lambda^{(\ell)}\|\lambda^{(\ell)})\le\varepsilon_\pi+\Ag e^{-c\Delta^2}.
\]
\end{lemma}
Comparing $P_\ell$ and $P$ and using Lemma~\ref{lem:sk-cross} to eliminate small cross terms, and using the good initialization of the regression terms, gives the following decoupling to the intuitive direction
\begin{lemma}[Group decoupling]\label{lem:sk-decouple}
Under the assumptions of Theorem~\ref{thm:global},
\[
P\ \ge \sum_{\ell\in[m]}\pi^*_\ell P_\ell - \Ag e^{-c\Delta^2}.
\]
\end{lemma}
Chaining Lemmas~\ref{lem:sk-decouple}, \ref{lem:sk-entropy} and \ref{lem:sk-lossKL} and absorbing the error terms into $c\varepsilon\le c\cL$ gives
\begin{lemma}[Descent direction]\label{lem:sk-descent}
Under the assumptions of Theorem~\ref{thm:global},
\[
P \ge c\ \cL(\bpi,\bmu,\btheta).
\]
\end{lemma}

Since $P\le\|\nabla_{\bmu,\btheta}\cL\|_F\,B$ when
$U\le B^2/2$ by Cauchy--Schwarz, Lemma~\ref{lem:sk-descent} gives $\|\nabla_{\bmu,\btheta}\cL\|_F\ge c\cL/B$. Two further ingredients close the argument (Appendix~\ref{sec:appB}). Local smoothness (Lemma~\ref{lem:smooth}) gives $\cL^{(t+1)}\le\cL^{(t)}-\tfrac\eta2\|\nabla\cL\|^2$, which with gradient dominance yields $1/\cL^{(t)}\ge1/\cL^{(0)}+c\eta t/B^2$, hence $T\lesssim B^2/(\eta\varepsilon)$. The same projection bound shows that $U$ is non-increasing, and since the expert part of the projection is a nonnegative quadratic form (Proposition~\ref{prop:quadform}), $U_\theta$ drifts by only $O(\eta L_0)$ over the run, which the choice of $\eta$ keeps inside $\Bth^2$.

\section{Experiments}
\label{sec:experiments}

Three ingredients of the analysis are not settled by the theorem: whether the width $n\gtrsim m\log m$ is needed, whether the least-squares warm start is needed, and what free mixing weights buy compared with the balancing that deployed MoEs impose. We test each. Data come from the model \eqref{eq:truth} and we run finite-sample gradient EM (Algorithm~\ref{alg:main} with sample averages) for at most $T=10{,}000$ updates on $N=10{,}000$ points. A run succeeds if the joint KL between the truth and the fitted model is at most $0.02$, measured on the training sample in Section~\ref{sec:exp-scaling} and on the held-out points otherwise. For pruning (Section~\ref{sec:exp-balance}) we report the held-out KL of the conditional density of $y$ given $\bx$, the quantity a deployed MoE is judged on. Full details are in Appendix~\ref{sec:expdetails}.

\subsection{Scaling with the number of experts}\label{sec:exp-scaling}

\begin{figure}[t]
    \centering
    \includegraphics[width=.55\textwidth]{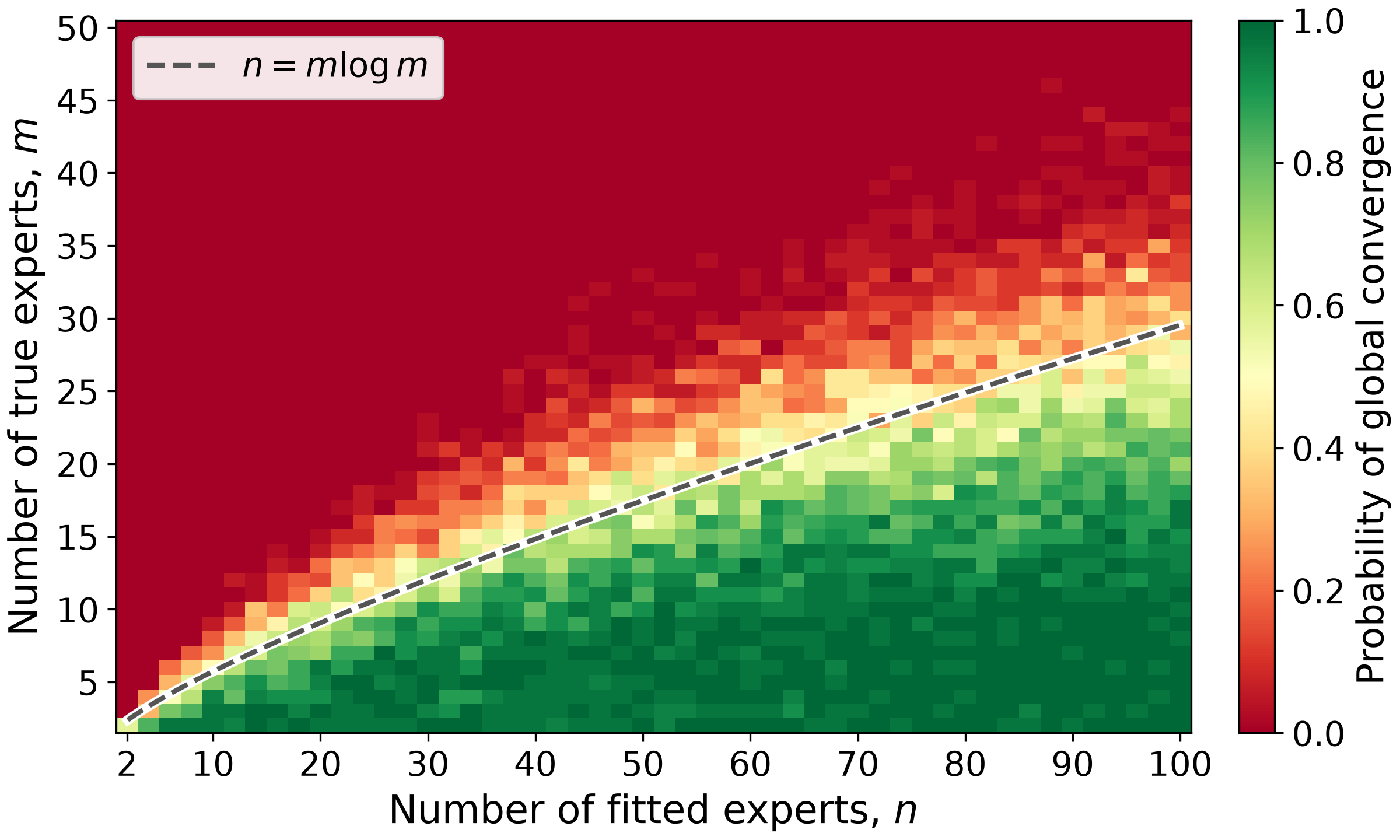}
    \caption{Probability of global convergence across parameterization regimes,
    with the LS warm start ($35$ trials per cell). The dashed line is the reference curve $n=m\log m$.}
    \label{fig:heatmap}
\end{figure}

We run a sweep to test where the failure boundary of $n$ versus $m$ lies. The question is whether the transition follows the diagonal $n=m$ (over-parameterization unnecessary) or the curve $n\asymp m\log m$ (the coverage mechanism of Lemma~\ref{lem:init}). Varying $m=2,\dots,50$ and $n=2,4,\dots,100$, Figure~\ref{fig:heatmap} shows a sharp transition along the reference curve $n \asymp m\log m$, with success near $n=m$ remaining low. The theorem only shows that $n\gtrsim m\log m$ suffices; the sharpness of the empirical transition along that curve, and the low success near $n=m$ for large $m$, suggest that logarithmic over-provisioning is also necessary for gradient EM from random gates.

\subsection{Necessity of the least-squares warm start}

Since modern MoE architectures train their experts and gates from random initialization, we compare LS initialization against $\btheta^{(0)}_i\sim N(0,I_d)$ with everything else held fixed. We sweep the fitted width $n\in\{4,8,\dots,24\}$ at weak ($\Delta=2$) and strong ($\Delta=12$) gate separation. By Lemma~\ref{lem:init} the warm start is most accurate when $\Delta$ is large, so $\Delta=12$ is where it should help most, and $\Delta=2$ is a stress test outside the theory.

\begin{figure}[t]
    \centering
    \includegraphics[width=.9\textwidth]{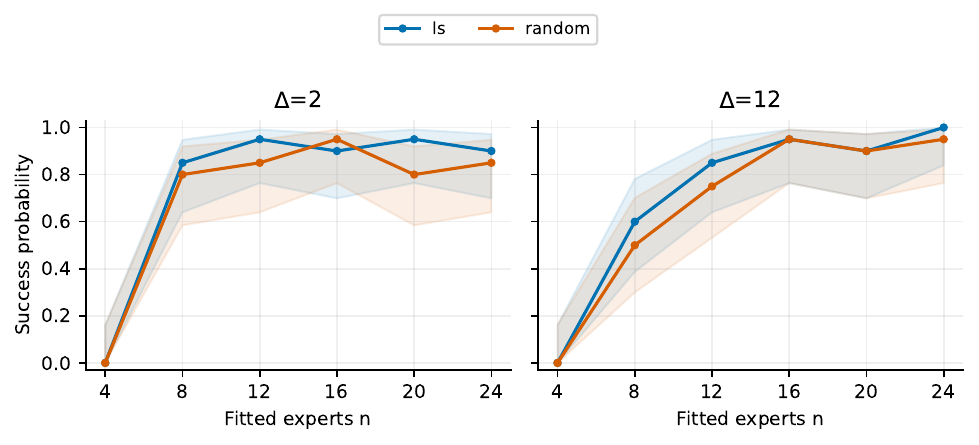}
    \caption{Success probability versus fitted width for LS and random expert
    initialization. \figleft\ Weak gate separation. \figright\ Strong gate
    separation. Shaded regions are confidence intervals over the $20$ seeds.}
    \label{fig:warm-start-ablation}
\end{figure}

In both panels of Figure~\ref{fig:warm-start-ablation} the two initializations fail at $n=4<m$, where the truth is not representable, and succeed together once the model is moderately over-parameterized. This indicates that the most important aspect is sufficient coverage by the initialized cluster centers $\bmu_i$. Once a gate lands near each true center, the experts follow regardless of where they started. Strong separation is in fact harder at small $n$ ($n=8$): a gate cannot cross a wide gap, so coverage at initialization is decisive, whereas for $\Delta=2$ an uncovered cluster can still attract a gate.

\subsection{Mixing weights, balancing, and effective experts}\label{sec:exp-balance}
Following the discussion in Section~\ref{sec:related}, we investigate how load balancing affects the benefits of freely learned mixing coefficients. We fit $n=16$ experts at $\Delta=8$, comparing unregularized training with the addition of $\lambda\KL(\bar{\boldsymbol\rho}\|\mathrm{Unif}_n)$ to the loss, where $\bar{\boldsymbol\rho}$ denotes average input-only router usage and $\lambda\in[0,1]$. Mixing coefficients remain trainable in both cases; the penalty encourages uniform usage across fitted experts. We consider uniform and skewed true mixture weights, specified in Appendix~\ref{sec:expdetails}.

\begin{figure}[t]
    \centering
    \includegraphics[width=.65\textwidth]{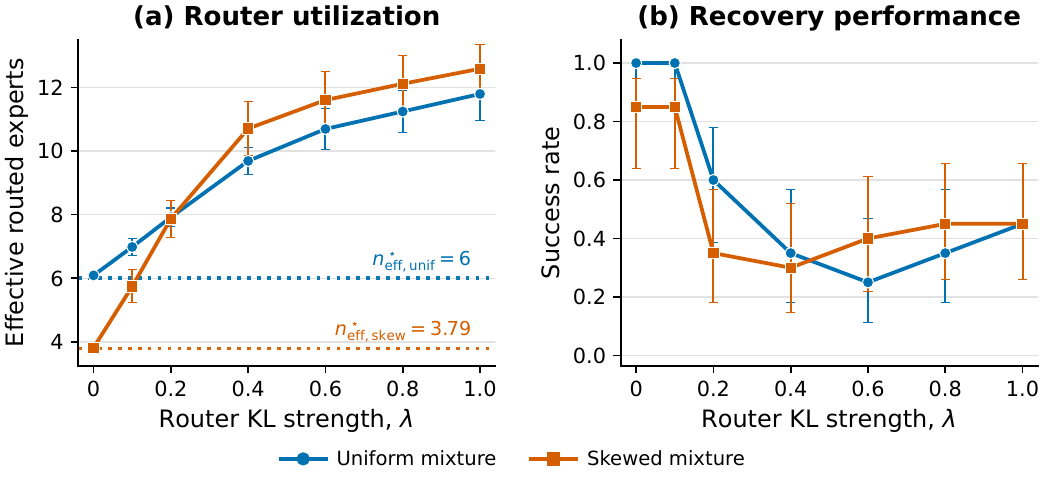}
    \caption{Effect of router load-balancing regularization on expert
    utilization and recovery. \textbf{(a)} Effective number of routed experts,
    $n_{\mathrm{eff}}$, against the router-KL strength $\lambda$ for uniform and
    skewed ground-truth mixture weights. Dotted lines show the corresponding
    oracle expert counts. \textbf{(b)} Recovery success against $\lambda$ for
    the same two ground truths. Error bars are $95\%$ confidence intervals.}
    \label{fig:balance-eff}
\end{figure}

We measure utilization by the effective expert count $n_{\mathrm{eff}}=\big(\sum_{i=1}^n\bar\rho_i^2\big)^{-1}$ of Section~\ref{sec:related}, where $\rho_i(\bx)=P(i\mid\bx)$ is the responsibility $\psi_i$ with $y$ marginalized out and $\bar\rho_i$ is its average over the held-out inputs. For large $\Delta$ the routing mass of a point from cluster $\ell$ falls almost entirely on $S_\ell$, so if a single expert carries each group, $\bar{\bm\rho}$ reproduces $\bpi^*$ and $n_{\mathrm{eff}}$ equals $n^*_{\mathrm{eff}}:=\big(\sum_{\ell=1}^m(\pi^*_\ell)^2\big)^{-1}$. For a uniform prior this is $n^*_{\mathrm{eff}}=(m\cdot m^{-2})^{-1}=m$ and $3.794$ for our skewed one. Free weights recover both almost exactly, at $6.09$ and $3.82$ (Figure~\ref{fig:balance-eff}). Raising $\lambda$ increases $n_{\mathrm{eff}}$ monotonically (Figure~\ref{fig:balance-eff}(a)), while the success rate drops from $1.00$ and $0.85$ for uniform and skewed mixtures, respectively, to between $0.25$ and $0.45$ for $\lambda\ge0.4$ (Figure~\ref{fig:balance-eff}(b)). Mild balancing increases effective utilization before reducing observed fitting success: at $\lambda=0.1$, the empirical success rates are unchanged, while $n_{\mathrm{eff}}$ rises to approximately $7.0$ and $5.4$, respectively. Higher utilization is therefore not a proxy for better training.  Encouraging uniform usage across fitted experts can reduce convergence success under the tested optimization budget, even when the true mixture weights are uniform.
\begin{figure}[htbp]
    \centering
    \includegraphics[width=.65\textwidth]{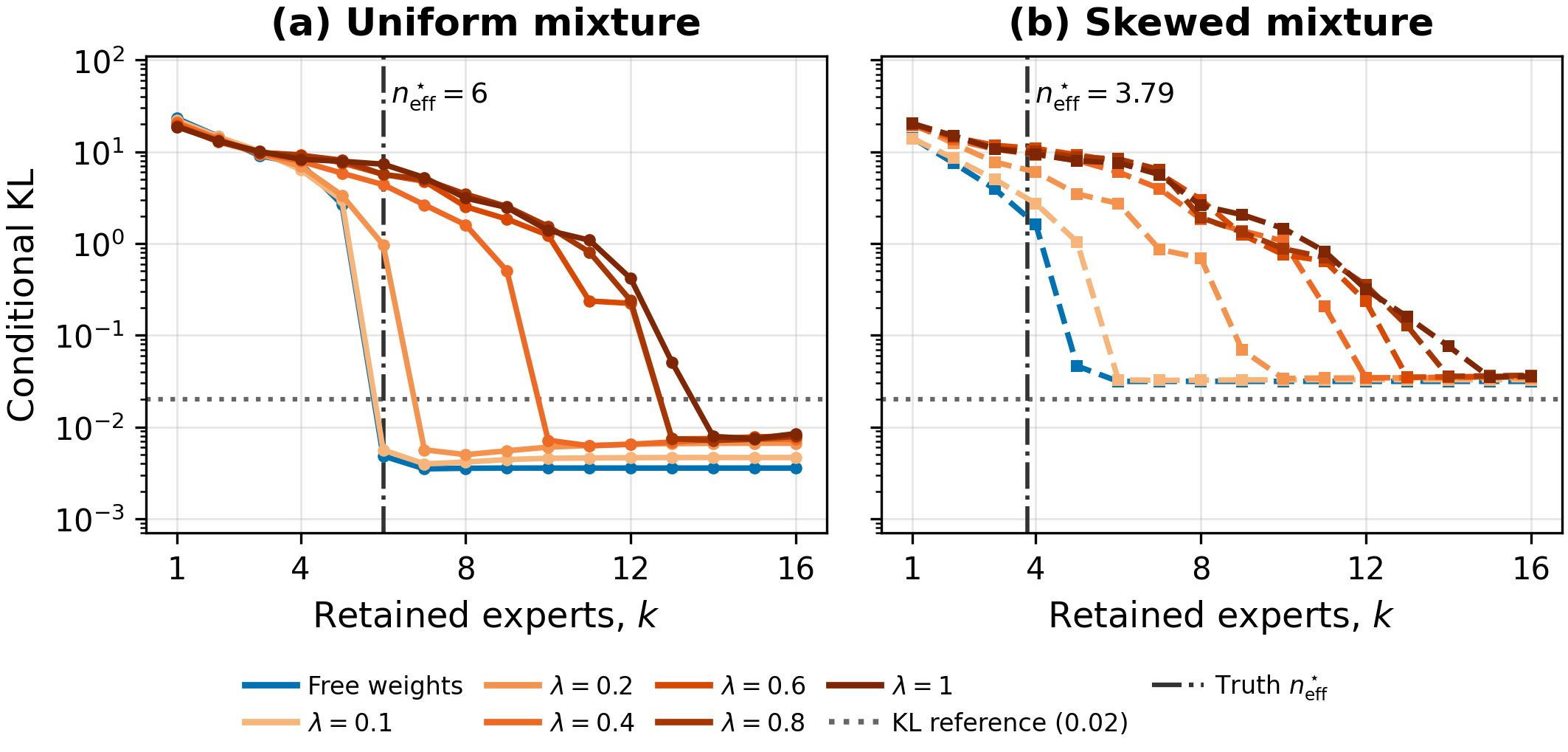}

    \caption{Effect of global expert pruning. Mean held-out conditional KL over all $20$ seeds, including unsuccessful training runs, is evaluated after retaining the $k$ experts with the largest learned mixing weights, renormalized without retraining, for \textbf{(a)} uniform and \textbf{(b)} skewed ground-truth mixture weights. The dotted line at $0.02$ is the joint-KL threshold used to define training success.}
    \label{fig:pruning-kl}
\end{figure}

The same effect decides whether over-parameterization pays off at inference. We prune to the $k$ experts with the largest learned mixing weights after the MoE has trained. Free-weight models reach their accuracy floor once $k\approx m$, while strongly balanced ones need nearly all the fitted experts (Figure~\ref{fig:pruning-kl}). Over-provisioning is what makes optimization succeed (Figure~\ref{fig:heatmap}); whether it also costs inference depends on whether the surplus can be discarded afterward, and that is decided by the router. A balancing penalty does the opposite: it drives $n_{\mathrm{eff}}$ towards $n$ by construction and leaves a more evenly used router with nothing to prune.

\section{Conclusion and Future Work}
\label{sec:conclusion}

We showed that gradient EM, equivalently gradient descent on the population KL loss, provably learns a Gaussian-gated mixture of $m$ linear experts from random gates and least-squares experts, provided the fitted width is $n\gtrsim m\log m$ and the gate centers are well separated. The experiments add two observations. First, the least-squares warm start is not needed in practice: randomly initialized experts converge almost as often once the gates cover the clusters. Second, free mixing weights concentrate routing on about $m$ experts, so the trained model can be pruned to the true width, whereas load-balancing penalties spread routing over most of the $n$ experts and leave little to prune.

Three questions follow. Can the least-squares warm start be removed from the analysis, as the experiments suggest? Furthermore, can the strong separation condition be relaxed as suggested in the same experiment? And does $n_{\mathrm{eff}}$ provably track $n^*_{\mathrm{eff}}$ at the end of training, as Figure~\ref{fig:balance-eff} indicates? Both experiments point to the same bottleneck: whether the initial gates cover the true clusters. Characterizing when gates are learnable from scratch, and connecting the optimization view taken here with the statistical rates of \citet{ho2022moe,nguyen2024gaussiangated}, is the natural next step.

\paragraph{Limitations.} Our guarantee is for population gradient EM; a full sample-based analysis, as \citet{zhou2026} give for Gaussian mixtures, remains to be done. The model has identity covariances, unit noise and linear experts, and the gate is induced by Gaussian clusters rather than a free softmax router. The separation required by Assumption~\ref{ass:sep} grows with the fitted width, which we believe to be an artifact of the potential $U$, and the step size must scale with $\varepsilon^2$, giving a $1/\varepsilon^3$ iteration count. Finally, the experts must be warm-started by least squares, although the experiments indicate that this is not needed in practice.

\bibliography{references}
\bibliographystyle{authoryear}

\appendix
\section{Notation, Constants, and Preliminaries}\label{sec:appA}

This appendix contains the proofs of all statements in the body of the paper.
Throughout, $c,C>0$ are universal constants whose value may change from line to
line, $a\lesssim b$ means $a\le Cb$, and
$\Ag=\poly\big(d,m,n,1/\pimin,1+B,1+\Dmax,1+\Dth\big)$
denotes a polynomial prefactor of universal degree which may be enlarged from
line to line. A single accuracy parameter $\varepsilon$ is used throughout: it is
simultaneously the target accuracy of Theorem~\ref{thm:global} and the threshold
below which the error terms of the analysis are absorbed. The Gaussian
observation noise is always written $\epsilon$, as in the data model of the main text.

\subsection{Notation}\label{sec:appnot}

Recall $\bz=(\bx,y)$, the truth \eqref{eq:truth}, the model \eqref{eq:model}, the
Voronoi groups $S_\ell$ and the potentials \eqref{eq:potentials}. We write
\begin{gather*}
\vphi_i(\bz):=\vphi(\bz;\bmu_i,\btheta_i),\qquad
\vphi^*_\ell(\bz):=\vphi(\bz;\bmu^*_\ell,\btheta^*_\ell),\\
\psi_i(\bz):=\frac{\pi_i\vphi_i(\bz)}{p(\bz)},\qquad
r_i(\bz):=y-\bx^\top\btheta_i .
\end{gather*}
for the components, the posterior (membership) weights and the residual of expert
$i$. Sampling from cluster $j$ means
\begin{equation}\label{eq:clustersample}
\bx=\bmu^*_j+\bm Z,\qquad y=\bx^\top\btheta^*_j+\epsilon,\qquad
(\bm Z,\epsilon)\sim N(0,I_{d+1}),
\end{equation}
and we write $\E_j$ for the corresponding expectation, so that
$\E_{p_*}=\sum_j\pi^*_j\E_j$. Relative to cluster $j$ we use the
recentered parameters
\begin{equation}\label{eq:recenterd}
\bh^{(j)}_k:=\bmu_k-\bmu^*_j,\qquad
\beta^{(j)}_k:=\btheta_k-\btheta^*_j,\qquad
a^{(j)}_k(\bx):=\bx^\top\beta^{(j)}_k,\qquad\text{so that } r_k=\epsilon-a^{(j)}_k .
\end{equation}
The superscript $(j)$ is dropped whenever the cluster is clear from the context.
Finally, for each group we use the normalized within-group mixture and its
posterior,
\begin{equation}\label{eq:withingroup}
\begin{gathered}
\lambda^{(\ell)}_i:=\frac{\pi_i}{\hat\pi_\ell},\qquad
q_\ell:=\sum_{i\in S_\ell}\lambda^{(\ell)}_i\vphi_i,\qquad
\psi^{(\ell)}_i:=\frac{\lambda^{(\ell)}_i\vphi_i}{q_\ell},\\[2pt]
\bar\lambda^{(\ell)}_i:=\E_\ell\big[\psi^{(\ell)}_i\big],\qquad
\cL_\ell:=\KL(\vphi^*_\ell\|q_\ell).
\end{gathered}
\end{equation}

\subsection{Constants and the sharp separation condition}\label{sec:appconst}

Fix $\delta\in(0,1)$ and a target accuracy $\varepsilon$. With
$\rho_0:=C\sqrt{d+\log(4n/\delta)}$ and $R:=\rho_0+2\sqrt d$ as in
Algorithm~\ref{alg:main}, set
\begin{equation}\label{eq:LB}
L_0:=\log n+2(d+1),\qquad B^2:=4n\rho_0^2 ,
\end{equation}
so that, on the event of Lemma~\ref{lem:init-app}, $\cL^{(0)}\le L_0$ and
$U(\bmu^{(0)},\btheta^{(0)})\le B^2/2$: item~2 of that lemma gives
$U_\mu(\bmu^{(0)})\le n\rho_0^2=B^2/4$, and \eqref{eq:U0budget} gives
$U_\theta(\bmu^{(0)},\btheta^{(0)})\le\Bth^2/4\le\tfrac14$. Since $n\ge1$ and
$C\ge1$ in the definition of $\rho_0$, we also have $B=2\sqrt n\,\rho_0\ge2$ and
$L_0\ge1$. Define the
\emph{expert-error budget}, the smoothness constant and the step-size and
weight-tolerance constants by
\begin{gather}
\Bth:=c_\theta\min\Big\{\frac{\varepsilon}{B},\ \frac1{(1+\Dth)\Lam}\Big\},
\label{eq:Btheta}\\[2pt]
A:=C(1+\Lam)^2\Big(1+d+B^2+\log\frac1{\pimin}+\frac1{\pimin}\Big),
\qquad
A_\eta:=\max\Big\{A,\ \frac{L_0}{\Bth^2}\Big\},
\qquad
A_\pi:=\frac{\Ag}{c},
\label{eq:Aconst}
\end{gather}
with $c_\theta$ a small universal constant. Note that $A_\eta$ depends on the target accuracy through $\Bth$: for small $\varepsilon$ we have $\Bth=c_\theta\varepsilon/B$, hence $A_\eta=L_0B^2/(c_\theta^2\varepsilon^2)$ and the admissible step size scales as $\varepsilon^2$. Since $B^2$ appears additively inside
$A$ with a large constant, and since
$B^2=4n\rho_0^2\ge4C^2\big(d+\log(4n/\delta)\big)\ge\log n+2(d+1)=L_0$ for
$n\ge1$ and $C\ge1$, we always have
\begin{equation}\label{eq:AgeL}
A\ \ge\ B^2\ \ge\ L_0 ,
\end{equation}
a fact used repeatedly to bound $\eta$-dependent quantities. The proofs use the
following sharp form of Assumption~\ref{ass:sep}:
\begin{equation}\label{eq:sepD1}
\Delta\ \ge\ C_{\mathrm{sep}}\Big(B+\sqrt{\log\Ag}
+\sqrt{\tfrac{L_0+\log(1/\pimin)}{\pimin}}
+\sqrt{\log\tfrac1{\varepsilon}}+\sqrt{\log\tfrac1{\Bth}}\Big),
\end{equation}
for a large universal constant $C_{\mathrm{sep}}\ge32$. Every term of
\eqref{eq:sepD1} is of the form $\sqrt{\poly}$ or $\sqrt{\log\poly}$, since $\Ag$
and $1/\Bth$ are polynomial in the problem parameters; Assumption~\ref{ass:sep}
is the readable form of this requirement, and the term
$\sqrt{\log(1/\varepsilon)}$ is what forces the restriction
$\varepsilon\ge e^{-c\,dn}$. Condition \eqref{eq:sepD1} is used in exactly two
ways: through $\Delta\ge C_{\mathrm{sep}}B$, which drives
Lemma~\ref{lem:constants}, and through the implication
\begin{equation}\label{eq:sepuse}
\Ag\,e^{-c\Delta^2}\ \le\ c\,\varepsilon\,\Bth^2/B^2 ,
\end{equation}
which follows by squaring \eqref{eq:sepD1} and taking $C_{\mathrm{sep}}$ large:
the terms $\sqrt{\log\Ag}$, $\sqrt{\log(1/\varepsilon)}$ and
$\sqrt{\log(1/\Bth)}$ contribute exactly the factors $\Ag^{-1}$, $\varepsilon$
and $\Bth^{2}$ on the right-hand side, and $B\le\Ag$.

Two standing consequences of the potential bounds will be used without comment:
\begin{equation}\label{eq:Bconseq}
U\le\tfrac{B^2}{2}\ \Longrightarrow\ \norm{\bmu_i-\bmu^*_\ell}\le B\ \ (i\in S_\ell),
\qquad
U_\theta\le\Bth^2\ \Longrightarrow\ \norm{\btheta_i-\btheta^*_\ell}\le\Bth\ \ (i\in S_\ell).
\end{equation}
Finally we abbreviate
\begin{equation}\label{eq:kappa}
\kappa:=2\Bth(\Dth+2\Bth).
\end{equation}

\begin{lemma}[Elementary consequences of the choice of constants]\label{lem:constants}
Assume Assumption~\ref{ass:sep} (in particular $m\ge2$) and
\eqref{eq:Btheta}--\eqref{eq:sepD1}. Then
\begin{enumerate}[nosep,label=(\roman*)]
\item $B\le\Delta\le2\Dmax$ and $\Lam+B^2\le5\Lam$;
\item $\Bth\le c_\theta\le1$ and $\Bth^2\Lam\le1$;
\item $\kappa(\Lam+B^2)\le c_0:=20\,c_\theta$;
\item $B<\Delta/4$ and $\Delta>2(B+1)$.
\end{enumerate}
\end{lemma}

\begin{proof}
(i) Condition \eqref{eq:sepD1} gives $\Delta\ge C_{\mathrm{sep}}B\ge B$. In the
working coordinates fixed in the main text, every true center satisfies
$\norm{\bmu^*_\ell}\le\Dmax$, so
$\Delta\le\norm{\bmu^*_\ell-\bmu^*_j}\le2\Dmax$ for any $\ell\ne j$. Hence
$B^2\le4\Dmax^2\le4\Lam$ because $\Lam=d+\Dmax^2\ge\Dmax^2$, and adding $\Lam$
gives $\Lam+B^2\le5\Lam$.

(ii) By \eqref{eq:Btheta}, $\Bth\le c_\theta/\big((1+\Dth)\Lam\big)\le c_\theta\le1$,
since $\Dth\ge0$ and $\Lam\ge d\ge1$. The same bound multiplied by $\Lam$ gives
\begin{equation}\label{eq:BthLam}
\Bth\Lam\ \le\ \frac{c_\theta}{1+\Dth}\ \le\ c_\theta ,
\end{equation}
so $\Bth^2\Lam=\Bth\cdot(\Bth\Lam)\le1\cdot c_\theta\le1$.

(iii) By (i) and \eqref{eq:BthLam},
\[
\kappa(\Lam+B^2)\ \le\ 2\Bth(\Dth+2\Bth)\cdot5\Lam
\ =\ 10\,(\Bth\Lam)(\Dth+2\Bth)
\ \le\ 10\,\frac{c_\theta}{1+\Dth}\,(\Dth+2)
\ \le\ 20\,c_\theta ,
\]
where we used $\Bth\le1$ in $\Dth+2\Bth\le\Dth+2$ and then
$\Dth+2\le2(1+\Dth)$.

(iv) $B\le\Delta/C_{\mathrm{sep}}\le\Delta/32<\Delta/4$. Since $B\ge2$,
$\Delta>2(B+1)$.
\end{proof}

\subsection{Gaussian moments}\label{sec:appmom}

\begin{lemma}[Moment bounds]\label{lem:moments}
For every $\ell\in[m]$ and every $\bm v\in\R^d$,
\begin{gather*}
\E_\ell\norm{\bx}^2\le2\Lam,\qquad
\E_\ell\norm{\bx}^4\le C\Lam^2,\\
\E_\ell\big[(\bx^\top\bm v)^2\big]\le2\Lam\norm{\bm v}^2,\qquad
\E_\ell\big[(\bx^\top\bm v)^4\big]\le C\Lam^2\norm{\bm v}^4 .
\end{gather*}
\end{lemma}

\begin{proof}
Write $\bx=\bmu^*_\ell+\bm Z$ with $\bm Z\sim N(0,I_d)$. Then
$\E_\ell\norm{\bx}^2=\norm{\bmu^*_\ell}^2+d\le\Dmax^2+d=\Lam$. For the fourth
moment, $(a+b)^4\le8a^4+8b^4$ and $\E\norm{\bm Z}^4=d(d+2)\le3d^2$ give
$\E_\ell\norm{\bx}^4\le8\Dmax^4+24d^2\le32\Lam^2$.
For the projections, $\bx^\top\bm v\sim N(\ip{\bmu^*_\ell,\bm v},\norm{\bm v}^2)$,
so its second moment is
$\ip{\bmu^*_\ell,\bm v}^2+\norm{\bm v}^2\le(\Dmax^2+1)\norm{\bm v}^2\le2\Lam\norm{\bm v}^2$,
and for $W\sim N(\varrho,s^2)$ one has $\E W^4=\varrho^4+6\varrho^2s^2+3s^4$, whence
$\E_\ell(\bx^\top\bm v)^4\le\big(\Dmax^4+6\Dmax^2+3\big)\norm{\bm v}^4\le C\Lam^2\norm{\bm v}^4$.
\end{proof}

\subsection{Stein's lemma and the gradient}\label{sec:appstein}

\begin{lemma}[Stein's lemma]\label{lem:stein}
Let $\bm w\sim N(\bm a,\sigma^2I_k)$ and let $g:\R^k\to\R$ be differentiable. Then
$\E\big[g(\bm w)(\bm w-\bm a)\big]=\sigma^2\,\E\big[\nabla g(\bm w)\big]$ when the expectations are well-defined.
Equivalently, for a differentiable function $F:\R^k\to\R^k$,
$\E\ip{\bm w-\bm a,F(\bm w)}=\sigma^2\E\big[\nabla\!\cdot\!F(\bm w)\big]$.
\end{lemma}

We apply Lemma~\ref{lem:stein} to the pair $(\bm Z,\epsilon)\sim N(0,I_{d+1})$ of
\eqref{eq:clustersample}. Every function to which it is applied below is a
posterior weight, a residual, or a product of these with a polynomial; all such
functions are smooth and of at most polynomial growth after the change of
variables, so the integrability hypotheses hold throughout and we do not repeat
this verification.

\begin{claim}[Gradient EM is gradient descent on $\cL$; restatement of Claim~\ref{claim:gradEM}]\label{claim:gradEM-app}
For any fixed $(\bpi',\bmu',\btheta')$,
$\nabla_{\bmu,\btheta}Q(\bpi,\bmu,\btheta\mid\bpi',\bmu',\btheta')\big|_{(\bpi',\bmu',\btheta')}
=-\nabla_{\bmu,\btheta}\cL(\bpi',\bmu',\btheta')$, and
\begin{equation}\label{eq:grad}
\nabla_{\bmu_i}\cL=\E_{p_*}\big[\psi_i(\bz)(\bmu_i-\bx)\big],\qquad
\nabla_{\btheta_i}\cL=-\E_{p_*}\big[\psi_i(\bz)\,r_i(\bz)\,\bx\big],\qquad
\nabla_{\pi_i}\cL=-\E_{p_*}\Big[\frac{\vphi_i}{p}\Big].
\end{equation}
\end{claim}

\begin{proof}
By \eqref{eq:loss}, $\cL=\E_{p_*}[-\log p]+\text{const}$, where the constant is
the (parameter-free) negative entropy of $p_*$. Since
$p=\sum_k\pi_k\vphi_k$ and only the $i$th summand depends on
$(\pi_i,\bmu_i,\btheta_i)$, the log-derivative identity gives
\[
\nabla_{\bmu_i}\cL=-\E_{p_*}\Big[\frac{\pi_i\nabla_{\bmu_i}\vphi_i}{p}\Big]
=-\E_{p_*}\Big[\frac{\pi_i\vphi_i}{p}\,\nabla_{\bmu_i}\log\vphi_i\Big]
=-\E_{p_*}\big[\psi_i(\bx-\bmu_i)\big],
\]
because $\log\vphi_i(\bz)=\text{const}-\tfrac12\norm{\bx-\bmu_i}^2-\tfrac12 r_i(\bz)^2$
has $\nabla_{\bmu_i}\log\vphi_i=\bx-\bmu_i$ and
$\nabla_{\btheta_i}\log\vphi_i=r_i\bx$; the same computation with
$\nabla_{\btheta_i}$ and with $\partial_{\pi_i}\vphi$ gives the two remaining
identities in \eqref{eq:grad}. In $Q$ the posterior $\psi'$ is frozen, so
$\nabla_{\bmu_i}Q=\E_{p_*}[\psi'_i(\bx-\bmu_i)]$ and
$\nabla_{\btheta_i}Q=\E_{p_*}[\psi'_ir_i\bx]$; evaluating at
$(\bmu,\btheta)=(\bmu',\btheta')$ makes $\psi'=\psi$ and yields exactly
$-\nabla_{\bmu,\btheta}\cL$.
\end{proof}

\begin{claim}[Stein form of the gradient]\label{claim:steinform}
Fix a data cluster $j$, use the recentered parameters \eqref{eq:recenterd}, and write
the posterior averages, in which the sums run over all $n$ components with the global posterior $\psi_k$,
\[
\begin{aligned}
\bar{\bh}^{(j)} &:= \sum_{k\in[n]}\psi_k\bh_k^{(j)}, & \bar{\bm w}^{(j)} &:= \sum_{k\in[n]}\psi_kr_k\beta_k^{(j)}, \\
\bar\beta^{(j)} &:= \sum_{k\in[n]}\psi_k\beta_k^{(j)}, & \bar a^{(j)} &:= \sum_{k\in[n]}\psi_ka_k^{(j)}=\bx^\top\bar\beta^{(j)} .
\end{aligned}
\]
Then
\begin{align}
\nabla_{\bmu_i}\cL&=\sum_j\pi^*_j\,\E_j\Big[\psi_i\big(\bar{\bh}^{(j)}+\bar{\bm w}^{(j)}-r_i\beta^{(j)}_i\big)\Big],\label{eq:steinmu}\\
\nabla_{\btheta_i}\cL&=\sum_j\pi^*_j\,\E_j\Big[\psi_i\,\bx\,\bx^\top\bar\beta^{(j)}\Big].\label{eq:steintheta}
\end{align}
\end{claim}

\begin{proof}
Fix $j$ and drop the superscript. Under cluster $j$ we compare each
$\log(\pi_k\vphi_k)$ with $\log\vphi^*_j$. Expanding
$\norm{\bx-\bmu_k}^2=\norm{\bx-\bmu^*_j}^2-2\ip{\bx-\bmu^*_j,\bh_k}+\norm{\bh_k}^2$
handles the $\bx$ part, and substituting $y=\bx^\top\btheta^*_j+\epsilon$ gives
$r_k=\epsilon-a_k$, hence
$-\tfrac12r_k^2+\tfrac12\epsilon^2=\epsilon a_k-\tfrac12a_k^2$
for the $y$ part. Therefore
$\log(\pi_k\vphi_k)-\log\vphi^*_j=\log\pi_k+f_k$ with
\begin{equation}\label{eq:fk}
f_k(\bz)=\ip{\bh_k,\bx-\bmu^*_j}-\tfrac12\norm{\bh_k}^2+\epsilon a_k-\tfrac12a_k^2,
\qquad \nabla_{\bx}f_k=\bh_k+r_k\beta_k,\qquad \partial_\epsilon f_k=a_k,
\end{equation}
where $\nabla_{\bx}\big(\epsilon a_k-\tfrac12a_k^2\big)=(\epsilon-a_k)\beta_k=r_k\beta_k$.
Since $\psi_i=e^{\log\pi_i+f_i}\big/\sum_ke^{\log\pi_k+f_k}$ is a softmax in the
$f_k$, the softmax Jacobian gives
\begin{equation}\label{eq:gradpsi}
\nabla_{\bx}\psi_i=\psi_i\Big(\nabla_{\bx}f_i-\sum_k\psi_k\nabla_{\bx}f_k\Big)
=\psi_i\big(\bh_i+r_i\beta_i-\bar{\bh}-\bar{\bm w}\big),
\qquad
\partial_\epsilon\psi_i=\psi_i(a_i-\bar a).
\end{equation}
For \eqref{eq:steinmu}, split
$\E_j[\psi_i(\bmu_i-\bx)]=\E_j[\psi_i]\bh_i-\E_j[\psi_i(\bx-\bmu^*_j)]$ and apply
Lemma~\ref{lem:stein} in $\bm Z=\bx-\bmu^*_j$ componentwise, so that
$\E_j[\psi_i(\bx-\bmu^*_j)]=\E_j[\nabla_{\bx}\psi_i]$; inserting
\eqref{eq:gradpsi} cancels the terms $\psi_i\bh_i$ and adds
$\psi_i(\bar{\bh}+\bar{\bm w}-r_i\beta_i)$, which is \eqref{eq:steinmu}.
For \eqref{eq:steintheta}, write $-\E_j[\psi_ir_i\bx]=-\E_j[\psi_i\epsilon\bx]+\E_j[\psi_ia_i\bx]$
and apply Lemma~\ref{lem:stein} in $\epsilon$ with $\bx$ held fixed:
$\E_j[\psi_i\epsilon\bx]=\E_j[\bx\,\partial_\epsilon\psi_i]=\E_j[\psi_i(a_i-\bar a)\bx]$.
The terms $\psi_ia_i\bx$ cancel and
$-\E_j[\psi_ir_i\bx]=\E_j[\psi_i\bar a\,\bx]=\E_j[\psi_i\bx\bx^\top\bar\beta]$.
\end{proof}

\subsection{The weight subproblem and the dual gap}\label{sec:dualgap}

For fixed $(\bmu,\btheta)$ the map $\bpi\mapsto\cL(\bpi,\bmu,\btheta)$ is convex
on the simplex $\cP$, since $p$ is linear in $\bpi$ and $-\log$ is convex. Its suboptimality is measured by the simplex dual gap
\begin{equation}\label{eq:gap}
G_\pi(\bpi,\bmu,\btheta):=\max_{\tilde\bpi\in\cP}\ip{\nabla_{\bpi}\cL,\bpi-\tilde\bpi}
=\max_{i}\big[-(\nabla_{\pi_i}\cL+1)\big],
\end{equation}
where the second equality uses
$\sum_i\pi_i\nabla_{\pi_i}\cL=-\E_{p_*}[p/p]=-1$ and the fact that a linear
function on $\cP$ is maximized at a vertex. A weight update is
\emph{$\varepsilon_\pi$-optimal} if
$G_\pi(\bpi^+,\bmu,\btheta)\le\varepsilon_\pi$ and
$\cL(\bpi^+,\bmu,\btheta)\le\cL(\bpi,\bmu,\btheta)$. Both properties hold for the
exact minimizer, and can be enforced for any monotone convex solver, e.g.\ the EM
fixed point $\pi_i\leftarrow\E_{p_*}\psi_i$, which is loss non-increasing and
converges to the minimizer.

\begin{lemma}[Consequences of the dual gap]\label{lem:dualgap}
Fix $(\bmu,\btheta)$ and let $\bpi\in\cP$ satisfy
$G_\pi(\bpi,\bmu,\btheta)\le\varepsilon_\pi$. Then
\[
\cL(\bpi,\bmu,\btheta)\le\inf_{\tilde\bpi\in\cP}\cL(\tilde\bpi,\bmu,\btheta)+\varepsilon_\pi
\qquad\text{and}\qquad
\sum_{i=1}^n\big|\pi_i-\E_{p_*}\psi_i\big|\le2\varepsilon_\pi .
\]
\end{lemma}

\begin{proof}
Write $g_i:=\nabla_{\pi_i}\cL$ and $\tilde{g}_i:=g_i+1$, so that $\tilde{g}_i\ge-\varepsilon_\pi$
for every $i$ by \eqref{eq:gap}. Convexity of $\cL$ in $\bpi$ gives, for any
$\tilde\bpi\in\cP$,
$\cL(\bpi)-\cL(\tilde\bpi)\le\ip{\nabla_{\bpi}\cL,\bpi-\tilde\bpi}\le\varepsilon_\pi$,
which is the first claim. For the second, $\sum_i\pi_i\tilde{g}_i=\sum_i\pi_ig_i+1=0$, so
the positive and negative parts of $(\tilde g_i)_i$ have equal $\bpi$-mass; each is at most
$\varepsilon_\pi$ because $\pi_i\tilde{g}_i^-\le\varepsilon_\pi\pi_i$ summed over $i$ is
at most $\varepsilon_\pi$. Hence $\sum_i\pi_i|\tilde{g}_i|\le2\varepsilon_\pi$, and the
claim follows from
$\pi_i\tilde{g}_i=\pi_i+\pi_i\nabla_{\pi_i}\cL=\pi_i-\E_{p_*}[\pi_i\vphi_i/p]=\pi_i-\E_{p_*}\psi_i$.
\end{proof}
\section{Initialization and Global Convergence}\label{sec:appB}

Throughout this appendix $S_\ell,\hat\pi_\ell,U_\mu,U_\theta,U$ are as in
\eqref{eq:potentials}, evaluated at the current iterate, and we freely use
\eqref{eq:Bconseq}, \eqref{eq:kappa} and Lemma~\ref{lem:constants}.

\subsection{Initialization: proof of Lemma~\ref{lem:init}}\label{sec:proofinit}

\begin{lemma}[Initialization; formal version of Lemma~\ref{lem:init}]\label{lem:init-app}
Let Assumption~\ref{ass:sep} hold, fix $\delta\in(0,1)$, suppose
$n\ge\frac{2}{\pimin}\log\frac{2m}{\delta}$, and run Algorithm~\ref{alg:main}
with $\rho_0=C\sqrt{d+\log(4n/\delta)}$ and $R=\rho_0+2\sqrt d$; assume
$\Delta\ge4(R+\rho_0)$, which is implied by \eqref{eq:sepD1}. Write $J_i$ for the
cluster from which $\bmu^{(0)}_i$ was drawn. Then with probability at least
$1-\delta$:
\begin{enumerate}[nosep]
\item for each $\ell\in[m]$ there is an $i$ with $J_i=\ell$ and
      $\norm{\bmu^{(0)}_i-\bmu^*_\ell}\le\sqrt{2d}$;
\item for every $i$, $\norm{\bmu^{(0)}_i-\bmu^*_{J_i}}\le\rho_0$; consequently
      $i\in S_{J_i}$ and $U_\mu(\bmu^{(0)})\le n\rho_0^2$;
\item \textup{(least-squares warm start)} for every $i$,
\[
\norm{\btheta^{(0)}_i-\btheta^*_{J_i}}\le\frac{C\Lam\,\Dth}{\pimin}\,e^{-\Delta^2/16},
\qquad\text{hence}\qquad
U_\theta(\bmu^{(0)},\btheta^{(0)})\le n\Big(\frac{C\Lam\Dth}{\pimin}\Big)^{\!2}e^{-\Delta^2/8};
\]
\item if in addition $\Lam\,U_\theta(\bmu^{(0)},\btheta^{(0)})\le1$, then
      $\cL^{(0)}\le\log n+\tfrac32(d+1)\le L_0$.
\end{enumerate}
\end{lemma}

\begin{proof}
Write the gate initialization as: draw $J_i\in[m]$ with probabilities $\bpi^*$ and
set $\bmu^{(0)}_i=\bmu^*_{J_i}+\bm Z_i$ with $\bm Z_i\sim N(0,I_d)$,
independently over $i$. This is exactly sampling $\bmu^{(0)}_i$ from the
$\bx$-marginal $p_*^{\bx}$.

\paragraph{Item 1 (coupon collector).}
Let $p:=\Pr\{\norm{\bm Z}^2\le2d\}$. Since $\E\norm{\bm Z}^2=d$, Markov's
inequality gives $\Pr\{\norm{\bm Z}^2>2d\}\le\frac12$, so $p\ge\frac12$. The
events $\{J_i=\ell\}$ and $\{\norm{\bm Z_i}^2\le2d\}$ are independent, so a
single draw succeeds for cluster $\ell$ with probability $\pi^*_\ell p$, and by
independence across $i$ the probability that no draw succeeds is
\[
(1-\pi^*_\ell p)^n\ \le\ \Big(1-\tfrac{\pimin}{2}\Big)^{\!n}\ \le\ e^{-n\pimin/2}\ \le\ \frac{\delta}{2m},
\]
using $1-u\le e^{-u}$ and $n\ge\frac2{\pimin}\log\frac{2m}\delta$. A union bound
over the $m$ clusters bounds the failure probability by $\delta/2$. On the
complementary event, $\norm{\bmu^{(0)}_i-\bmu^*_\ell}=\norm{\bm Z_i}\le\sqrt{2d}$
for the selected $i$.

\paragraph{Item 2 (concentration).}
The map $\bm Z\mapsto\norm{\bm Z}$ is $1$-Lipschitz, so Gaussian concentration
gives $\Pr\{\norm{\bm Z_i}>\sqrt d+t\}\le e^{-t^2/2}$. Taking
$t=\sqrt{2\log(4n/\delta)}$ makes each failure probability $\delta/(4n)$, and a
union bound over $i$ leaves probability at least $1-\delta/4$ that
\[
\norm{\bm Z_i}\ \le\ \sqrt d+\sqrt{2\log(4n/\delta)}\ \le\ \sqrt3\,\sqrt{d+\log(4n/\delta)}\ \le\ \rho_0
\qquad\text{for all }i ,
\]
where the middle step is Cauchy--Schwarz, $\sqrt a+\sqrt{2b}\le\sqrt3\sqrt{a+b}$,
and the last step holds for $C\ge\sqrt3$ in the definition of $\rho_0$. By
Lemma~\ref{lem:constants}(iv) and $\Delta\ge4(R+\rho_0)>4\rho_0$ we get
$\norm{\bmu^{(0)}_i-\bmu^*_{J_i}}\le\rho_0<\Delta/4<\Delta/2$; since distinct true
centers are at distance at least $\Delta$, $\bmu^*_{J_i}$ is the strictly nearest
true center, i.e.\ $i\in S_{J_i}$. Summing,
$U_\mu(\bmu^{(0)})\le\sum_i\norm{\bm Z_i}^2\le n\rho_0^2$. Items~1 and 2 hold
simultaneously with probability at least $1-\tfrac34\delta\ge1-\delta$.

\paragraph{Item 3 (least-squares warm start).}
Fix $i$, write $\ell=J_i$ and $w(\bx):=\ind\{\norm{\bx-\bmu^{(0)}_i}\le R\}$.
Since $R=\rho_0+2\sqrt d$ and $\norm{\bmu^{(0)}_i-\bmu^*_\ell}\le\rho_0$, the
triangle inequality gives the nesting
\[
B(\bmu^*_\ell,2\sqrt d)\ \subset\ B(\bmu^{(0)}_i,R)\ \subset\ B(\bmu^*_\ell,R+\rho_0),
\]
and, for $j\ne\ell$, every $\bx$ in the window satisfies
$\norm{\bx-\bmu^*_j}\ge\Delta-(R+\rho_0)\ge\tfrac34\Delta$ because
$\Delta\ge4(R+\rho_0)$.

The objective in \eqref{eq:lsinit} is the convex quadratic
$\btheta\mapsto\btheta^\top M\btheta-2\bm b^\top\btheta+\text{const}$ with
\begin{gather*}
M:=\E_{p_*}[w(\bx)\bx\bx^\top]=\sum_{j}\pi^*_jM_j,\qquad
M_j:=\E_j[w(\bx)\bx\bx^\top],\\
\bm b:=\E_{p_*}[w(\bx)\bx y]=\sum_j\pi^*_jM_j\btheta^*_j ,
\end{gather*}
the last identity because under cluster $j$ we have $y=\bx^\top\btheta^*_j+\epsilon$
with $\E[\epsilon\mid\bx]=0$, so the noise cross term vanishes. Once $M\succ0$ is
established, $\btheta^{(0)}_i=M^{-1}\bm b$, and subtracting
$\btheta^*_\ell=M^{-1}M\btheta^*_\ell=M^{-1}\sum_j\pi^*_jM_j\btheta^*_\ell$ gives
\begin{equation}\label{eq:lserr}
\btheta^{(0)}_i-\btheta^*_\ell=M^{-1}\sum_{j\ne\ell}\pi^*_jM_j(\btheta^*_j-\btheta^*_\ell).
\end{equation}

\emph{Lower bound on $M$.} By the inner inclusion,
$w(\bx)\ge\ind\{\norm{\bm Z}\le2\sqrt d\}$ for $\bx=\bmu^*_\ell+\bm Z$. For a unit
vector $\bm u$, expanding
$\ip{\bm u,\bmu^*_\ell+\bm Z}^2=\ip{\bm u,\bmu^*_\ell}^2+2\ip{\bm u,\bmu^*_\ell}\ip{\bm u,\bm Z}+\ip{\bm u,\bm Z}^2$
and noting that the domain $\{\norm{\bm Z}\le2\sqrt d\}$ is invariant under
$\bm Z\mapsto-\bm Z$ (so the odd cross term integrates to $0$) yields
\begin{align*}
\bm u^\top M_\ell\bm u\ &\ge\ \ip{\bm u,\bmu^*_\ell}^2\Pr\{\norm{\bm Z}\le2\sqrt d\}
+\E\big[\ind\{\norm{\bm Z}\le2\sqrt d\}\ip{\bm u,\bm Z}^2\big]\\
&\ge\ \E\big[\ind\{\norm{\bm Z}\le2\sqrt d\}Z_1^2\big]=:c_1 ,
\end{align*}
where the last equality uses rotational invariance to take $\bm u=\bm e_1$. By the
law of total expectation and Cauchy--Schwarz with $\E Z_1^4=3$,
\[
c_1=\E[Z_1^2]-\E\big[Z_1^2\ind\{\norm{\bm Z}>2\sqrt d\}\big]
\ \ge\ 1-\sqrt{3\Pr\{\norm{\bm Z}>2\sqrt d\}} .
\]
For $d\ge6$, $\Pr\{\norm{\bm Z}>2\sqrt d\}\le e^{-d/2}$ gives $c_1>\tfrac12$; for
$d\le5$, $c_1$ is the integral of a strictly positive function over a set of
positive measure, hence a fixed positive number. In either case $c_1$ is a
universal positive constant and, since all $M_j\succeq0$,
$M\succeq\pi^*_\ell M_\ell\succeq c_1\pimin I_d$.

\emph{Upper bound on $M_j$, $j\ne\ell$.} Since $M_j\succeq0$, its operator norm is
at most its trace, so by Cauchy--Schwarz, $w^2=w$ and Lemma~\ref{lem:moments},
\begin{align*}
\norm{M_j}_{\op}&\le\E_j\big[w(\bx)\norm{\bx}^2\big]
\le\sqrt{\E_j\norm{\bx}^4}\,\sqrt{\Pr\nolimits_j\{\norm{\bx-\bmu^{(0)}_i}\le R\}}\\
&\le C\Lam\sqrt{\Pr\{\norm{\bm Z}\ge\tfrac34\Delta\}}\ \le\ C\Lam e^{-\Delta^2/16}.
\end{align*}
The last step uses $\Pr\{\norm{\bm Z}\ge\sqrt d+t\}\le e^{-t^2/2}$ with
$t=\tfrac34\Delta-\sqrt d\ge\tfrac12\Delta$, which holds because
$\Delta\ge4R\ge8\sqrt d$.

Substituting both bounds into \eqref{eq:lserr}, using
$\norm{\btheta^*_j-\btheta^*_\ell}\le\Dth$ and $\sum_{j\ne\ell}\pi^*_j\le1$,
\[
\norm{\btheta^{(0)}_i-\btheta^*_\ell}\ \le\ \frac{1}{c_1\pimin}\cdot C\Lam e^{-\Delta^2/16}\cdot\Dth ,
\]
and squaring and summing over the $n$ components bounds $U_\theta$ as claimed.

\paragraph{Item 4 (initial loss).}
Fix $\ell$ and let $i$ be the component supplied by item~1, so that
$\norm{\bmu^{(0)}_i-\bmu^*_\ell}^2\le2d$ and
$\norm{\btheta^{(0)}_i-\btheta^*_\ell}^2\le U_\theta^{(0)}$. Since
$p_*\le(2\pi)^{-(d+1)/2}$ pointwise (a mixture of unit-covariance Gaussian
densities) and $p^{(0)}\ge\frac1n\vphi_i$, the normalizing constants cancel and
\[
\log\frac{p_*(\bz)}{p^{(0)}(\bz)}\ \le\ \log n+\tfrac12\norm{\bx-\bmu^{(0)}_i}^2+\tfrac12r_i(\bz)^2 .
\]
Under cluster $\ell$, the bias--variance decomposition gives
$\E_\ell\norm{\bx-\bmu^{(0)}_i}^2=d+\norm{\bmu^*_\ell-\bmu^{(0)}_i}^2\le3d$, and
$r_i=\epsilon+\bx^\top(\btheta^*_\ell-\btheta^{(0)}_i)$ with $\E[\epsilon\mid\bx]=0$
gives, by Lemma~\ref{lem:moments},
\[
\E_\ell r_i^2=1+\E_\ell\big[(\bx^\top(\btheta^*_\ell-\btheta^{(0)}_i))^2\big]
\le1+2\Lam\,U_\theta^{(0)}\le3 .
\]
Averaging over $\ell$ with weights $\pi^*_\ell$ gives
$\cL^{(0)}\le\log n+\tfrac32 d+\tfrac32$, which is at most $L_0$ by
\eqref{eq:LB}.
\end{proof}

\begin{remark}\label{rem:initbudget}
Combining item~3 with \eqref{eq:sepuse} and $\varepsilon\le c\,\pimin\le1$, $B\ge2$,
\begin{equation}\label{eq:U0budget}
U_\theta(\bmu^{(0)},\btheta^{(0)})\ \le\ \Ag e^{-\Delta^2/8}\ \le\ \frac{\Bth^2}{4},
\end{equation}
so the hypothesis $\Lam U_\theta^{(0)}\le1$ of item~4 holds by
Lemma~\ref{lem:constants}(ii), and the expert potential starts at a quarter of
its budget.
\end{remark}

\subsection{The global-convergence theorem}\label{sec:proofthm}

\begin{theorem}[Global convergence of gradient EM; formal version of Theorem~\ref{thm:global}]\label{thm:global-app}
Let Assumption~\ref{ass:sep} and the separation condition \eqref{eq:sepD1} hold,
work on the event of Lemma~\ref{lem:init-app}, and let
$L_0,B,\Bth,A,A_\eta,A_\pi$ be as in \eqref{eq:LB}--\eqref{eq:Aconst}. Fix a
target accuracy $\varepsilon$ with $e^{-c\,dn}\le\varepsilon\le c\,\pimin$.
Assume each weight update is loss non-increasing and that every post-weight point
satisfies the dual-gap condition of Appendix~\ref{sec:dualgap},
\begin{equation}\label{eq:dualcond}
G_\pi\big(\bpi^{(t+1)},\bmu^{(t)},\btheta^{(t)}\big)\le\varepsilon_\pi,
\qquad \varepsilon_\pi\le\frac{\varepsilon}{A_\pi}=\frac{c\,\varepsilon}{\Ag}.
\end{equation}
If $\eta\le c/A_\eta$, then
\[
\cL\big(\bpi^{(t)},\bmu^{(t)},\btheta^{(t)}\big)\le\varepsilon
\qquad\text{for some }t\le T:=\frac{CB^2}{\eta\varepsilon},
\]
and for every $t\le T$ before the target is reached,
\[
U(\bmu^{(t)},\btheta^{(t)})\le\frac{B^2}{2},\qquad
U_\theta(\bmu^{(t)},\btheta^{(t)})\le\Bth^2,\qquad\text{and the groups }S_\ell\text{ do not change.}
\]
\end{theorem}

For small $\varepsilon$ we have $A_\eta=L_0B^2/(c_\theta^2\varepsilon^2)$ by \eqref{eq:Btheta}--\eqref{eq:Aconst}, so the largest admissible step size $\eta=c/A_\eta$ gives $T=CB^2A_\eta/(c\,\varepsilon)=O(L_0B^4/\varepsilon^3)$: the iteration count is polynomial in $(d,m,n,1/\pimin,\log(1/\delta),\Dmax,\Dth)$ and of order $1/\varepsilon^{3}$. The lower restriction
$\varepsilon\ge e^{-c\,dn}$ is exactly what makes \eqref{eq:sepD1} compatible
with Assumption~\ref{ass:sep}, whose separation is at least of order $\sqrt{dn}$.

The proof combines a one-step statement (Proposition~\ref{prop:onestep-app}
below) with the following control on the accumulated drift of the expert
potential.

\begin{lemma}[Expert-potential drift along the trajectory]\label{lem:thetadrift}
Let $(\bpi^{(s)},\bmu^{(s)},\btheta^{(s)})_{s\le t+1}$ be iterates of
Algorithm~\ref{alg:main} such that the hypotheses of
Proposition~\ref{prop:onestep-app} hold at every post-weight point up to time
$t$. Then
\[
U_\theta^{(t+1)}\ \le\ U_\theta^{(0)}+2\eta(t+1)\Ag e^{-c\Delta^2}+2\eta\big(L_0-\cL^{(t+1)}\big).
\]
\end{lemma}

\begin{proof}
Write $\cL^{(s+\frac12)}:=\cL(\bpi^{(s+1)},\bmu^{(s)},\btheta^{(s)})$ for the
post-weight loss. Since the weight step does not move $(\bmu,\btheta)$ and is
loss non-increasing, $\cL^{(s+\frac12)}\le\cL^{(s)}$. The third inequality of
Proposition~\ref{prop:onestep-app} and the descent inequality
$\cL^{(s+1)}\le\cL^{(s+\frac12)}-\frac\eta2\norm{\nabla\cL^{(s+\frac12)}}_F^2$ of
Lemma~\ref{lem:smooth} (applicable under those hypotheses, as verified in
Appendix~\ref{sec:proofprop}) give, for every $s\le t$,
\begin{align*}
U_\theta^{(s+1)}-U_\theta^{(s)}
&\le2\eta\Ag e^{-c\Delta^2}+\eta^2\norm{\nabla\cL^{(s+\frac12)}}_F^2\\
&\le2\eta\Ag e^{-c\Delta^2}+2\eta\big(\cL^{(s+\frac12)}-\cL^{(s+1)}\big)
\ \le\ 2\eta\Ag e^{-c\Delta^2}+2\eta\big(\cL^{(s)}-\cL^{(s+1)}\big),
\end{align*}
where the middle step multiplies the descent inequality by $2\eta$. Summing over
$s=0,\dots,t$ telescopes both the left-hand side and the loss differences, and
$\cL^{(0)}\le L_0$ gives the claim. The potential $U_\theta$ is evaluated with
the groups $S_\ell$, which do not change along the trajectory by
Proposition~\ref{prop:invariance}, so the telescoping is legitimate: no term in
the sum is affected by a reassignment.
\end{proof}

\begin{proof}[Proof of Theorem~\ref{thm:global-app}]
Write $L_t:=\cL(\bpi^{(t)},\bmu^{(t)},\btheta^{(t)})$,
$L_{t+\frac12}:=\cL(\bpi^{(t+1)},\bmu^{(t)},\btheta^{(t)})$,
$U_t:=U(\bmu^{(t)},\btheta^{(t)})$ and $U_{\theta,t}$ likewise, and let $\tau$ be
the first time at which the loss is at most $\varepsilon$. We prove by induction
on $t<\tau$, $t\le T$, that
\begin{equation}\label{eq:induction}
L_t\le L_0,\qquad U_t\le\frac{B^2}{2},\qquad U_{\theta,t}\le\Bth^2,\qquad
S_\ell(\bmu^{(t)})=S_\ell(\bmu^{(0)})\ \ \forall\ell .
\end{equation}

\emph{Base case.} At $t=0$ the first two statements are the initialization event
together with the choices \eqref{eq:LB}, and the third is the sharper bound
$U_{\theta,0}\le\Bth^2/4$ of \eqref{eq:U0budget}.

\emph{Induction step.} Suppose \eqref{eq:induction} holds at time $t<\tau$. The
weight step leaves $(\bmu,\btheta)$ and hence $U,U_\theta,S_\ell$ unchanged, and
is loss non-increasing, so $L_{t+\frac12}\le L_t$. If $L_{t+\frac12}\le\varepsilon$
we are done. Otherwise $\varepsilon<L_{t+\frac12}\le L_0$, so
Proposition~\ref{prop:onestep-app} applies at the post-weight point and gives
$U_{t+1}\le U_t\le B^2/2$ and
$L_{t+1}\le L_{t+\frac12}-\frac{c\eta}{B^2}L_{t+\frac12}^2\le L_0$; the groups
are unchanged by Proposition~\ref{prop:invariance}. For the expert potential,
Lemma~\ref{lem:thetadrift} together with \eqref{eq:U0budget} gives
\[
U_{\theta,t+1}\ \le\ \frac{\Bth^2}{4}+2\eta(t+1)\Ag e^{-c\Delta^2}+2\eta L_0
\ \le\ \frac{\Bth^2}{4}+\frac{2CB^2}{\varepsilon}\Ag e^{-c\Delta^2}+\frac{\Bth^2}{4}
\ \le\ \Bth^2 ,
\]
where the middle step uses $t+1\le T=CB^2/(\eta\varepsilon)$, so that the factor
$\eta$ cancels, together with
$\eta\le c/A_\eta\le c\,\Bth^2/L_0\le\Bth^2/(8L_0)$ for $c\le\frac18$; and the
last step uses \eqref{eq:sepuse} in the form
$\Ag e^{-c\Delta^2}\le\varepsilon\Bth^2/(8CB^2)$. This completes the induction.

\emph{Recurrence.} We convert the per-step decrease into a recurrence for $L_t$.
If $L_{t+\frac12}\ge L_t/2$, then substituting this lower bound into the
quadratic term and $L_{t+\frac12}\le L_t$ into the linear term gives
$L_{t+1}\le L_t-\frac{c\eta}{4B^2}L_t^2$. If instead $L_{t+\frac12}<L_t/2$, then
$L_{t+1}\le L_{t+\frac12}<L_t/2\le L_t-\frac{c\eta}{B^2}L_t^2$, the last
inequality being equivalent to $\frac{c\eta}{B^2}L_t\le\frac12$, which holds
because $\eta\le c/A$, $L_t\le L_0\le A$ by \eqref{eq:AgeL} and $B^2\ge1$, so
that $\frac{c\eta L_t}{B^2}\le c^2\frac{L_0}{A}\frac1{B^2}\le c^2\le\frac12$.
Absorbing the factor $\tfrac14$ into $c$, in both cases
\[
L_{t+1}\le L_t-\frac{c\eta}{B^2}L_t^2
\qquad\Longrightarrow\qquad
\frac1{L_{t+1}}-\frac1{L_t}=\frac{L_t-L_{t+1}}{L_tL_{t+1}}\ \ge\ \frac{L_t-L_{t+1}}{L_t^2}\ \ge\ \frac{c\eta}{B^2},
\]
using $L_{t+1}\le L_t$ (if $L_{t+1}=0$ the target is already reached). Summing,
$1/L_t\ge1/L_0+c\eta t/B^2$, hence $L_t\le B^2/(c\eta t)\le\varepsilon$ as soon as
$t\ge B^2/(c\eta\varepsilon)$. Therefore $\tau\le T=CB^2/(\eta\varepsilon)$, which
is within the range of the induction, and \eqref{eq:induction} holds up to
$\tau$.
\end{proof}

\subsection{One global step}\label{sec:proofprop}

\begin{proposition}[One global step]\label{prop:onestep-app}
Under the assumptions of Theorem~\ref{thm:global-app}, consider a post-weight
point $(\bpi,\bmu,\btheta)$ with
\[
\varepsilon\le\cL:=\cL(\bpi,\bmu,\btheta)\le L_0,\qquad
U(\bmu,\btheta)\le\frac{B^2}2,\qquad U_\theta(\bmu,\btheta)\le\Bth^2,
\]
satisfying \eqref{eq:dualcond}, and let
$(\bmu^+,\btheta^+):=(\bmu,\btheta)-\eta\nabla_{\bmu,\btheta}\cL(\bpi,\bmu,\btheta)$.
If $\eta\le c/A$, then
\begin{align*}
U(\bmu^+,\btheta^+)&\le U(\bmu,\btheta),\qquad
\cL(\bpi,\bmu^+,\btheta^+)\le\cL-\frac{c\eta}{B^2}\cL^2,\\
U_\theta(\bmu^+,\btheta^+)&\le U_\theta(\bmu,\btheta)+2\eta\Ag e^{-c\Delta^2}
+\eta^2\norm{\nabla_{\bmu,\btheta}\cL(\bpi,\bmu,\btheta)}_F^2 .
\end{align*}
\end{proposition}

\begin{proof}
Write $g:=\nabla_{\bmu,\btheta}\cL(\bpi,\bmu,\btheta)=(g^\mu,g^\theta)$ and let
$P=P_\mu+P_\theta$ be the projection \eqref{eq:P} defined below.

\emph{Verification of the smoothness hypotheses.} By
Lemma~\ref{lem:groupweight}, $\hat\pi_\ell\ge\pi^*_\ell/2$ for every $\ell$, so
Lemma~\ref{lem:smooth} applies with $c_{\mathrm{cov}}=\tfrac12$. By
\eqref{eq:Bconseq}, $\norm{\bmu_i-\bmu^*_\ell}\le B$ and
$\norm{\btheta_i-\btheta^*_\ell}\le\Bth\le1$ for $i\in S_\ell$, hence
\[
\frac1{\hat\pi_\ell}\sum_{i\in S_\ell}\pi_i\big(\norm{\bmu_i-\bmu^*_\ell}^2+\norm{\btheta_i-\btheta^*_\ell}^2\big)\le B^2+\Bth^2\le2B^2 ,
\]
and the factor $2$ is absorbed into the constant $C$ inside $A$. Lemma~\ref{lem:smooth}
therefore gives
\begin{equation}\label{eq:smoothuse}
\norm{g}_F^2\le2A\cL,\qquad
\cL(\bpi,\bmu-\eta g^\mu,\btheta-\eta g^\theta)\le\cL-\frac\eta2\norm{g}_F^2
\qquad(\eta\le c/A).
\end{equation}

\emph{Total potential.} By Proposition~\ref{prop:invariance} the groups do not
change, so $U$ may be evaluated at the new point with the same assignment.
Expanding each squared distance,
$\norm{(\bmu_i-\bmu^*_\ell)-\eta g^\mu_i}^2=\norm{\bmu_i-\bmu^*_\ell}^2-2\eta\ip{g^\mu_i,\bmu_i-\bmu^*_\ell}+\eta^2\norm{g^\mu_i}^2$
and likewise for $\btheta$, and summing over $i\in S_\ell$ and $\ell$,
\[
U(\bmu^+,\btheta^+)=U-2\eta P+\eta^2\norm{g}_F^2
\ \le\ U-2c\eta\cL+2\eta^2A\cL\ =\ U-2\eta\cL\,(c-\eta A)\ \le\ U ,
\]
by Lemma~\ref{lem:descent-app} ($P\ge c\cL$), \eqref{eq:smoothuse} and
$\eta\le c/A$.

\emph{Loss decrease.} The projection is the inner product of $g$ with the
displacement vector, whose squared norm is exactly $U$, so Cauchy--Schwarz gives
$P\le\norm{g}_F\sqrt U\le\norm{g}_FB/\sqrt2$, i.e.\
$\norm{g}_F\ge\sqrt2\,P/B\ge c\,\cL/B$ by Lemma~\ref{lem:descent-app}. Inserting
this into the descent inequality of \eqref{eq:smoothuse},
\[
\cL(\bpi,\bmu^+,\btheta^+)\ \le\ \cL-\frac\eta2\norm{g}_F^2\ \le\ \cL-\frac{c\eta}{B^2}\cL^2 .
\]

\emph{Expert potential.} Exactly the same expansion restricted to the
$\btheta$-block gives
$U_\theta(\bmu^+,\btheta^+)=U_\theta-2\eta P_\theta+\eta^2\norm{g^\theta}_F^2$.
The second part of Lemma~\ref{lem:descent-app} bounds
$-2\eta P_\theta\le2\eta\Ag e^{-c\Delta^2}$, and
$\norm{g^\theta}_F^2\le\norm{g}_F^2$.
\end{proof}

\subsection{The descent direction}\label{sec:proofdescent}

Recall the projection onto the ``move every gate and every expert towards the
truth of its group'' direction,
\begin{equation}\label{eq:P}
P:=\sum_{\ell\in[m]}\sum_{i\in S_\ell}
\Big(\ip{\nabla_{\bmu_i}\cL,\bmu_i-\bmu^*_\ell}+\ip{\nabla_{\btheta_i}\cL,\btheta_i-\btheta^*_\ell}\Big)
=:P_\mu+P_\theta .
\end{equation}
For $i\in S_\ell$ define the integrands
\begin{equation}\label{eq:Ti}
T_i(\bz):=\ip{\bmu_i-\bx,\bmu_i-\bmu^*_\ell}-r_i(\bz)\,\bx^\top(\btheta_i-\btheta^*_\ell),
\qquad
T^\theta_i(\bz):=-r_i(\bz)\,\bx^\top(\btheta_i-\btheta^*_\ell),
\end{equation}
so that, by \eqref{eq:grad}, $P=\E_{p_*}\big[\sum_i\psi_iT_i\big]$ and
$P_\theta=\E_{p_*}\big[\sum_i\psi_iT^\theta_i\big]$. The \emph{one-cluster
projection} of cluster $\ell$ replaces the global posterior by the within-group
posterior and the mixture $p_*$ by the single component $\vphi^*_\ell$:
\begin{equation}\label{eq:Pell}
P_\ell:=\E_\ell\Big[\sum_{i\in S_\ell}\psi^{(\ell)}_i(\bz)\,T_i(\bz)\Big],
\qquad
P_{\ell,\theta}:=\E_\ell\Big[\sum_{i\in S_\ell}\psi^{(\ell)}_i(\bz)\,T^\theta_i(\bz)\Big].
\end{equation}
In the recentered coordinates of cluster $\ell$,
\begin{equation}\label{eq:recoord}
\begin{gathered}
\bx':=\bx-\bmu^*_\ell,\qquad y':=y-\bx^\top\btheta^*_\ell,\\[2pt]
\bh_i:=\bmu_i-\bmu^*_\ell,\qquad\beta_i:=\btheta_i-\btheta^*_\ell,\qquad
a_i:=\ip{\bx',\beta_i}+\ip{\bmu^*_\ell,\beta_i}=\bx^\top\beta_i ,
\end{gathered}
\end{equation}
one has $r_i=y'-a_i$ and
\begin{equation}\label{eq:Tirecenterd}
T_i=\norm{\bh_i}^2-\ip{\bx',\bh_i}+a_i^2-y'a_i,\qquad T^\theta_i=a_i^2-y'a_i ,
\end{equation}
because $\ip{\bmu_i-\bx,\bmu_i-\bmu^*_\ell}=\ip{\bh_i-\bx',\bh_i}$ and
$-r_i\bx^\top\beta_i=-(y'-a_i)a_i$. Thus $P_\ell$ is exactly the quantity $P$ of
the abstract one-cluster problem of Appendix~\ref{sec:onecluster}.

The following two statements are the formal versions of the group-decoupling and
descent lemmas stated in the body of the paper.

\begin{lemma}[Group decoupling; formal version of Lemma~\ref{lem:sk-decouple}]\label{lem:decouple}
Under the assumptions of Proposition~\ref{prop:onestep-app},
\[
P\ \ge\ \sum_{\ell\in[m]}\pi^*_\ell P_\ell-\Ag e^{-c\Delta^2},
\qquad
P_\theta\ \ge\ \sum_{\ell\in[m]}\pi^*_\ell P_{\ell,\theta}-\Ag e^{-c\Delta^2}.
\]
\end{lemma}

\begin{lemma}[Descent direction; formal version of Lemma~\ref{lem:sk-descent}]\label{lem:descent-app}
Under the assumptions of Proposition~\ref{prop:onestep-app},
$P\ge c\,\cL(\bpi,\bmu,\btheta)$ and $P_\theta\ge-\Ag e^{-c\Delta^2}$.
\end{lemma}

\begin{proof}[Proof of Lemma~\ref{lem:decouple}]
\emph{A polynomial envelope.} Fix a data cluster $j$ and any component $i$, with
$\ell=\ell(i)$, and recall $\epsilon=y-\bx^\top\btheta^*_j$. By
\eqref{eq:Bconseq} and the triangle inequality,
$\norm{\bx}\le\norm{\bx-\bmu^*_j}+\Dmax$,
$\norm{\bmu_i-\bx}\le\norm{\bx-\bmu^*_j}+2\Dmax+B$ and
$\abs{r_i}\le\abs{\epsilon}+\norm{\bx}(\Dth+1)$, since
$\norm{\btheta_i-\btheta^*_j}\le\Dth+\Bth\le\Dth+1$. Each of $T_i,T^\theta_i$ is a
product of at most three such factors, so
\begin{equation}\label{eq:envelope}
\abs{T_i(\bz)}+\abs{T^\theta_i(\bz)}\ \le\ W_j(\bz):=C(1+B)^2(1+\Dmax)^2(1+\Dth)\big(1+\norm{\bx-\bmu^*_j}+\abs{\epsilon}\big)^3 ,
\end{equation}
a weight of the form covered by Lemma~\ref{lem:cross} with $K=3$.

\emph{Reduction to same-group terms.} Split
$\E_{p_*}=\sum_j\pi^*_j\E_j$ and $\sum_i=\sum_{i\in S_j}+\sum_{i\notin S_j}$. For
the foreign components, \eqref{eq:envelope} and Lemma~\ref{lem:cross}(2) give
\[
\Big|\E_j\Big[\sum_{i\notin S_j}\psi_iT_i\Big]\Big|
\ \le\ \E_j\Big[\sum_{i\notin S_j}\psi_iW_j\Big]\ \le\ \Ag e^{-c\Delta^2}.
\]
For $i\in S_j$ we factor the global posterior through the within-group one:
\[
\psi_i=\frac{\pi_i\vphi_i}{p}=\frac{\hat\pi_jq_j}{p}\cdot\frac{\lambda^{(j)}_i\vphi_i}{q_j}=\rho_j\,\psi^{(j)}_i,
\qquad \rho_j=\frac{1}{1+\chi_j}\in(0,1],
\]
with $\chi_j$ as in Lemma~\ref{lem:cross}(3). Hence
\[
\E_j\Big[\sum_{i\in S_j}\psi_iT_i\Big]
=P_j-\E_j\Big[(1-\rho_j)\sum_{i\in S_j}\psi^{(j)}_iT_i\Big]
\ \ge\ P_j-\E_j\big[(1-\rho_j)W_j\big]\ \ge\ P_j-\Ag e^{-c\Delta^2},
\]
again by Lemma~\ref{lem:cross}(3). Multiplying by $\pi^*_j$, summing over $j$ and
using $\sum_j\pi^*_j=1$ (the factor $m$ and the constant $2$ are absorbed into
$\Ag$) proves the first inequality. Since $T^\theta_i$ obeys the same envelope
\eqref{eq:envelope}, the identical argument proves the second.
\end{proof}

\begin{proof}[Proof of Lemma~\ref{lem:descent-app}]
\emph{The expert part.} By Lemma~\ref{lem:decouple} and
Proposition~\ref{prop:quadform}, which states $P_{\ell,\theta}\ge0$ for every
$\ell$, we may drop the (nonnegative) sum and obtain
$P_\theta\ge-\Ag e^{-c\Delta^2}$.

\emph{One-cluster projection comparison.} Apply Lemma~\ref{lem:sk-entropy-app} with
$\tau:=c\varepsilon/B^2$, which lies in $(0,1]$ because
$\varepsilon\le c\pimin\le1\le B^2$. The resulting deficit is
\[
\frac{\tau}{2}B^2+6\Big(1+\frac1{2\tau}\Big)\Bth^2
=\frac{c\varepsilon}{2}+6\Bth^2+\frac{3B^2\Bth^2}{c\,\varepsilon}
\ \le\ \frac{c\varepsilon}{2}+\Big(6c_\theta^2+\frac{3c_\theta^2}{c}\Big)\varepsilon
\ \le\ c\,\varepsilon ,
\]
where we used $\Bth\le c_\theta\varepsilon/B$ from \eqref{eq:Btheta}, so that
$\Bth^2\le c_\theta^2\varepsilon^2/B^2\le c_\theta^2\varepsilon$, and then chose
$c_\theta$ small enough given $c$. Writing $\cL_\ell=\KL(\vphi^*_\ell\|q_\ell)$
and using the second part of Lemma~\ref{lem:sk-entropy-app},
\[
\sum_\ell\pi^*_\ell P_\ell
\ \ge\ 2\sum_\ell\pi^*_\ell\cL_\ell-2\sum_\ell\pi^*_\ell\KL(\bar\lambda^{(\ell)}\|\lambda^{(\ell)})-c\varepsilon
\ \ge\ 2\sum_\ell\pi^*_\ell\cL_\ell-2\varepsilon_\pi-\Ag e^{-c\Delta^2}-c\varepsilon .
\]

\emph{Comparison with the loss.} By Lemma~\ref{lem:lossKL-app},
$\sum_\ell\pi^*_\ell\cL_\ell\ge\cL-C\varepsilon_\pi-\Ag e^{-c\Delta^2}$.
Combining with Lemma~\ref{lem:decouple},
\[
P\ \ge\ 2\cL-C\varepsilon_\pi-\Ag e^{-c\Delta^2}-c\varepsilon\ \ge\ \cL ,
\]
because $\varepsilon_\pi\le c\Ag^{-1}\varepsilon$ by \eqref{eq:dualcond},
$\Ag e^{-c\Delta^2}\le c\varepsilon$ by \eqref{eq:sepuse}, and
$\cL\ge\varepsilon$, so that the three error terms together are at most
$c\varepsilon\le c\cL\le\cL$.
\end{proof}
\section{Technical Lemmas for the Global Convergence}\label{sec:appC}

The inputs to this appendix are the Gaussian ratio bounds of
Lemma~\ref{lem:ratio}, proved in Appendix~\ref{sec:appD}, which we restate in the
form used here. Under \eqref{eq:Bconseq}, \eqref{eq:kappa} and \eqref{eq:sepD1},
for every data cluster $j$, every $\ell\ne j$, $i\in S_\ell$, $k\in S_j$,
$r\ne j$, and every polynomial weight
$W_K(\bz)=(1+\norm{\bx-\bmu^*_j}+\abs{\epsilon})^K$ of fixed degree $K$,
\begin{equation}\label{eq:ratiorecap}
\begin{gathered}
\E_j\Big[\Big(\frac{\vphi_i}{\vphi_k}\Big)^{\!1/2}W_K\Big]\le\Ag e^{-c\Delta^2},
\qquad
\int\sqrt{\vphi^*_j}\;\vphi_i\,(\vphi^*_\ell)^{-1/2}\,d\bz\le\Ag e^{-c\Delta^2},\\[2pt]
\E_j\Big[\Big(\frac{\vphi^*_r}{\vphi^*_j}\Big)^{\!1/2}\Big]\le e^{-\Delta^2/8}.
\end{gathered}
\end{equation}

\subsection{Cross-group estimates}\label{sec:appcross}

\begin{lemma}[Cross-group estimates; formal version of Lemma~\ref{lem:sk-cross}]\label{lem:cross}
Under the assumptions of Proposition~\ref{prop:onestep-app} the following hold for every $\ell\ne j$, every
$i\in S_\ell$ and every polynomial weight
$W(\bz)=(1+\norm{\bx-\bmu^*_j}+\abs{\epsilon})^K$ of fixed degree $K$, with
$\epsilon=y-\bx^\top\btheta^*_j$:
\begin{enumerate}[nosep]
\item $\hat\pi_\ell\ge\exp\big(-C\frac{L_0+\log(1/\pimin)}{\pimin}\big)$ for every $\ell$;
\item $\E_j[\psi_i(\bz)W(\bz)]\le\Ag e^{-c\Delta^2}$, and consequently
      \[
      \E_\ell\Big[\sum_{i\notin S_\ell}\psi_iW\Big]+\sum_{j\ne\ell}\E_j\Big[\sum_{i\in S_\ell}\psi_iW\Big]\le\Ag e^{-c\Delta^2};
      \]
\item with $p_{-\ell}:=\sum_{r\ne\ell}\sum_{i\in S_r}\pi_i\vphi_i$,
      $\chi_\ell:=p_{-\ell}/(\hat\pi_\ell q_\ell)$ and
      $\rho_\ell:=1/(1+\chi_\ell)=\hat\pi_\ell q_\ell/p$,
      \[
      \E_\ell\big[\log(1+\chi_\ell)\big]+\E_\ell\big[(1-\rho_\ell)W\big]\le\Ag e^{-c\Delta^2};
      \]
\item with $\chi^*_\ell:=\sum_{r\ne\ell}\pi^*_r\vphi^*_r/(\pi^*_\ell\vphi^*_\ell)$,
      $\E_\ell[\log(1+\chi^*_\ell)]\le\Ag e^{-c\Delta^2}$.
\end{enumerate}
The quantities $\chi_\ell$ and $\chi^*_\ell$ are the interference ratios of the
model and of the truth: they factor $p=\hat\pi_\ell q_\ell(1+\chi_\ell)$
and $p_*=\pi^*_\ell\vphi^*_\ell(1+\chi^*_\ell)$ for data from cluster $\ell$.
\end{lemma}

\begin{proof}
If $m=1$ there is nothing to prove, so assume $m\ge2$.

\emph{Item 1 (lower bound on the group mass).}
Since $p_*\ge\pi^*_r\vphi^*_r$ pointwise for every $r$, for $i\in S_\ell$ and
data from cluster $\ell$,
\[
\E_\ell\Big[\frac{\vphi_i}{p_*}\Big]\le\frac1{\pi^*_\ell}\E_\ell\Big[\frac{\vphi_i}{\vphi^*_\ell}\Big]
=\frac1{\pi^*_\ell}\int\vphi_i\,d\bz=\frac1{\pi^*_\ell},
\]
because $\E_\ell$ integrates against $\vphi^*_\ell$, which cancels. For
$i\in S_r$ with $r\ne\ell$ we instead take the geometric mean of the two
pointwise bounds $p_*\ge\pi^*_\ell\vphi^*_\ell$ and $p_*\ge\pi^*_r\vphi^*_r$:
\[
\E_\ell\Big[\frac{\vphi_i}{p_*}\Big]
\le\frac1{\pimin}\E_\ell\Big[\Big(\frac{\vphi_i}{\vphi^*_\ell}\Big)^{\!1/2}\Big(\frac{\vphi_i}{\vphi^*_r}\Big)^{\!1/2}\Big]
=\frac1{\pimin}\int\sqrt{\vphi^*_\ell}\;\vphi_i\,(\vphi^*_r)^{-1/2}\,d\bz
\le\Ag e^{-c\Delta^2}
\]
by the second bound in \eqref{eq:ratiorecap}. Multiplying by $\pi_i$ and summing,
and noting that the components inside $S_\ell$ carry total mass $\hat\pi_\ell$,
\begin{equation}\label{eq:massupper}
\begin{gathered}
\E_\ell\Big[\frac p{p_*}\Big]\le\frac{\hat\pi_\ell}{\pi^*_\ell}+\Ag e^{-c\Delta^2},
\qquad\text{and in particular}\\[2pt]
\E_r\Big[\frac p{p_*}\Big]\le\frac1{\pimin}+\Ag e^{-c\Delta^2}\le\frac2{\pimin}
\quad\text{for every }r .
\end{gathered}
\end{equation}
In the other direction, since $-\log$ is convex, Jensen's inequality gives
\[
L_0\ \ge\ \cL=\sum_r\pi^*_r\E_r\Big[-\log\frac p{p_*}\Big]
\ \ge\ -\sum_r\pi^*_r\log\E_r\Big[\frac p{p_*}\Big]
\ \ge\ -\pi^*_\ell\log\E_\ell\Big[\frac p{p_*}\Big]-\log\frac2{\pimin},
\]
where the last step bounds the $r\ne\ell$ terms using the second display in
\eqref{eq:massupper} and $\sum_{r\ne\ell}\pi^*_r\le1$. Rearranging,
$\E_\ell[p/p_*]\ge\exp\big(-\frac{L_0+\log(2/\pimin)}{\pi^*_\ell}\big)$. Combining
with \eqref{eq:massupper} and $\pi^*_\ell\ge\pimin$,
\[
\hat\pi_\ell\ \ge\ \pimin\exp\Big(-\frac{L_0+\log(2/\pimin)}{\pimin}\Big)-\Ag e^{-c\Delta^2},
\]
and by the second and third terms of \eqref{eq:sepD1} the subtracted term is at most half the first, which is
absorbed by enlarging $C$ in the exponent.

\emph{Item 2 (posterior cross terms).}
Fix $j\ne\ell$ and $i\in S_\ell$. Since $\abs{S_j}\le n$, the pigeonhole
principle supplies $k\in S_j$ with $\pi_k\ge\hat\pi_j/n$. Because
$p\ge\pi_k\vphi_k$ and $\psi_i\le1$, and since $\min\{1,u\}\le\sqrt u$ for
$u\ge0$,
\[
\psi_i=\frac{\pi_i\vphi_i}{p}\le\min\Big\{1,\frac{\pi_i\vphi_i}{\pi_k\vphi_k}\Big\}
\le\Big(\frac{\pi_i}{\pi_k}\Big)^{\!1/2}\Big(\frac{\vphi_i}{\vphi_k}\Big)^{\!1/2}
\le\Big(\frac{n}{\hat\pi_j}\Big)^{\!1/2}\Big(\frac{\vphi_i}{\vphi_k}\Big)^{\!1/2}.
\]
By item~1, $(n/\hat\pi_j)^{1/2}\le\sqrt n\exp\big(C\frac{L_0+\log(1/\pimin)}{\pimin}\big)$,
a factor that \eqref{eq:sepD1} absorbs into $\Ag e^{-c\Delta^2}$ after halving
$c$, since $\Delta^2\gtrsim(L_0+\log(1/\pimin))/\pimin$. The first bound in
\eqref{eq:ratiorecap} then gives $\E_j[\psi_iW]\le\Ag e^{-c\Delta^2}$. Summing
over at most $n$ indices $i$ and $m$ clusters, and absorbing $nm$ into $\Ag$,
proves the second statement.

\emph{Item 3 (model interference ratio).}
For $\bz$ from cluster $\ell$ pick $k\in S_\ell$ with $\pi_k\ge\hat\pi_\ell/n$;
since $\hat\pi_\ell q_\ell\ge\pi_k\vphi_k$,
\[
\chi_\ell=\frac{p_{-\ell}}{\hat\pi_\ell q_\ell}\le\sum_{i\notin S_\ell}\frac{\pi_i\vphi_i}{\pi_k\vphi_k}.
\]
We use the elementary inequalities $\log(1+u)\le\sqrt u$ and
$\frac{u}{1+u}\le\sqrt u$ for $u\ge0$ (both follow from
$\sqrt u\ge\min\{u,1\}$, and $\tfrac{d}{du}\big(\sqrt u-\log(1+u)\big)=\frac{1+u-2\sqrt u}{2\sqrt u(1+u)}\ge0$),
together with the subadditivity $\sqrt{\sum_iu_i}\le\sum_i\sqrt{u_i}$. Since
$1-\rho_\ell=\chi_\ell/(1+\chi_\ell)$,
\[
\log(1+\chi_\ell)+(1-\rho_\ell)W\ \le\ (1+W)\sqrt{\chi_\ell}
\ \le\ (1+W)\sum_{i\notin S_\ell}\Big(\frac{\pi_i}{\pi_k}\Big)^{\!1/2}\Big(\frac{\vphi_i}{\vphi_k}\Big)^{\!1/2}.
\]
Taking $\E_\ell$ and applying the first bound in \eqref{eq:ratiorecap} (with data
cluster $\ell$, $i\in S_r$ for $r\ne\ell$, and $k\in S_\ell$), with the weight
ratio absorbed exactly as in item~2, proves item~3.

\emph{Item 4 (true interference ratio).}
By the same two inequalities,
\[
\log(1+\chi^*_\ell)\ \le\ \sqrt{\chi^*_\ell}\ \le\ \sum_{r\ne\ell}\Big(\frac{\pi^*_r}{\pi^*_\ell}\Big)^{\!1/2}\Big(\frac{\vphi^*_r}{\vphi^*_\ell}\Big)^{\!1/2},
\]
and the third bound in \eqref{eq:ratiorecap} gives
$\E_\ell\log(1+\chi^*_\ell)\le m\,\pimin^{-1/2}e^{-\Delta^2/8}\le\Ag e^{-c\Delta^2}$.
\end{proof}

\subsection{Group-weight control}\label{sec:appgw}

\begin{lemma}[Group-weight control]\label{lem:groupweight}
Under the assumptions of Proposition~\ref{prop:onestep-app},
\[
\sum_{\ell=1}^m\abs{\hat\pi_\ell-\pi^*_\ell}\le2\varepsilon_\pi+\Ag e^{-c\Delta^2},
\qquad\text{and hence}\qquad
\hat\pi_\ell\ge\frac{\pi^*_\ell}{2}\quad\forall\ell .
\]
\end{lemma}

\begin{proof}
By Lemma~\ref{lem:dualgap}, $\sum_i\abs{\pi_i-\E_{p_*}\psi_i}\le2\varepsilon_\pi$.
Writing $\Psi_\ell:=\sum_{i\in S_\ell}\psi_i$ and using that the $S_\ell$
partition $[n]$, the triangle inequality gives
$\sum_\ell\abs{\hat\pi_\ell-\E_{p_*}\Psi_\ell}\le\sum_i\abs{\pi_i-\E_{p_*}\psi_i}\le2\varepsilon_\pi$.
Next, splitting $\E_{p_*}=\sum_j\pi^*_j\E_j$ and using $\sum_{i=1}^n\psi_i=1$,
\[
\E_{p_*}\Psi_\ell=\pi^*_\ell\Big(1-\E_\ell\Big[\sum_{i\notin S_\ell}\psi_i\Big]\Big)+\sum_{j\ne\ell}\pi^*_j\E_j\Psi_\ell ,
\]
so that
$\sum_\ell\abs{\E_{p_*}\Psi_\ell-\pi^*_\ell}\le\sum_\ell\E_\ell\big[\sum_{i\notin S_\ell}\psi_i\big]+\sum_\ell\sum_{j\ne\ell}\E_j\big[\sum_{i\in S_\ell}\psi_i\big]\le\Ag e^{-c\Delta^2}$
by Lemma~\ref{lem:cross}(2) with $K=0$. A final triangle inequality proves the
first claim.

For the second, \eqref{eq:dualcond}, \eqref{eq:sepuse} and
$\varepsilon\le c\,\pimin$ give
$2\varepsilon_\pi+\Ag e^{-c\Delta^2}\le c\,\varepsilon\le\pimin/2$. Since every
summand is nonnegative, $\abs{\hat\pi_\ell-\pi^*_\ell}\le\pimin/2\le\pi^*_\ell/2$
for each $\ell$, whence $\hat\pi_\ell\ge\pi^*_\ell/2$.
\end{proof}

\subsection{One-cluster projection comparison}\label{sec:onecluster}

We first isolate an abstract one-cluster statement; the mixture-of-experts
application follows by the recentering \eqref{eq:recoord}.

Let $\nu$ be the standard Gaussian on $(\bx',y')\in\R^{d+1}$. Fix weights
$\lambda\in\cP^{N-1}$, vectors $\bh_i,\beta_i\in\R^d$ and scalars $c_i\in\R$,
define the affine regressions $a_i(\bx'):=\ip{\bx',\beta_i}+c_i$, and set
\begin{equation}\label{eq:oneclustermodel}
q(\bx',y'):=\sum_i\lambda_i\,\vphi(\bx';\bh_i)\,\vphi_1(y';a_i(\bx')),\qquad
\psi_i:=\frac{\lambda_i\vphi(\bx';\bh_i)\vphi_1(y';a_i)}{q},\qquad r_i:=y'-a_i .
\end{equation}
Write $\bar\lambda_i:=\E_\nu\psi_i$ and introduce the posterior means, variances
and cross-covariance
\begin{gather*}
\bar{\bh}:=\sum_i\psi_i\bh_i,\qquad \bar{\bm w}:=\sum_i\psi_ir_i\beta_i,\qquad \bar a:=\sum_i\psi_ia_i,\\
V_h:=\sum_i\psi_i\norm{\bh_i-\bar{\bh}}^2,\qquad
V_w:=\sum_i\psi_i\norm{r_i\beta_i-\bar{\bm w}}^2,\qquad
V_y:=\sum_i\psi_i(a_i-\bar a)^2,\\
\mathcal C:=\sum_i\psi_i\ip{r_i\beta_i-\bar{\bm w},\,\bh_i-\bar{\bh}},
\end{gather*}
and the one-cluster projection
\begin{equation}\label{eq:Pone}
P:=\E_\nu\Big[\sum_i\psi_i\big(\norm{\bh_i}^2-\ip{\bx',\bh_i}+a_i^2-y'a_i\big)\Big].
\end{equation}

\begin{remark}[{The recentering is measure preserving}]\label{rem:shear}
For cluster $\ell$ of the mixture of experts, \eqref{eq:oneclustermodel} is
$q_\ell$ written in the coordinates \eqref{eq:recoord}, with
$\lambda=\lambda^{(\ell)}$, $\bh_i=\bmu_i-\bmu^*_\ell$,
$\beta_i=\btheta_i-\btheta^*_\ell$ and $c_i=\ip{\bmu^*_\ell,\beta_i}$; indeed
$y-\bx^\top\btheta_i=y'-(\ip{\bx',\beta_i}+\ip{\bmu^*_\ell,\beta_i})=r_i$. The map
$(\bx,y)\mapsto(\bx',y')$ is the shift-and-shear
$(\bx,y)\mapsto(\bx-\bmu^*_\ell,\;y-\bx^\top\btheta^*_\ell)$, whose Jacobian matrix is
lower triangular with unit diagonal and hence has determinant $1$. Consequently
$\KL(\vphi^*_\ell\|q_\ell)=\KL(\nu\|q)$ and
$\E_\ell\psi^{(\ell)}_i=\E_\nu\psi_i$, and by \eqref{eq:Tirecenterd} the
projection \eqref{eq:Pone} coincides with $P_\ell$ of \eqref{eq:Pell}.
\end{remark}

\begin{lemma}[One-cluster projection comparison]\label{lem:onecluster}
In the setting \eqref{eq:oneclustermodel}, for every $\tau\in(0,1]$,
\[
P\ \ge\ 2\KL(\nu\|q)-2\KL(\bar\lambda\|\lambda)-\frac\tau2\,\E_\nu V_h-\Big(1+\frac1{2\tau}\Big)\E_\nu V_w .
\]
More precisely, one has the exact identity and the entropy bound
\begin{equation}\label{eq:exactP}
\begin{gathered}
P=2\KL(\nu\|q)-2\,\E_\nu\Big[\sum_i\psi_i\log\frac{\psi_i}{\lambda_i}\Big]+\E_\nu\big[V_h+V_y+\mathcal C\big],\\[2pt]
\E_\nu\Big[\sum_i\psi_i\log\frac{\psi_i}{\bar\lambda_i}\Big]\le\tfrac12\E_\nu\big[V_h+2\mathcal C+V_w+V_y\big].
\end{gathered}
\end{equation}
\end{lemma}

\begin{proof}
Indices with $\lambda_i=0$ are dropped. Write
\[
f_i:=\log\frac{\vphi(\bx';\bh_i)\vphi_1(y';a_i)}{\vphi(\bx';0)\vphi_1(y';0)}
=\ip{\bh_i,\bx'}-\tfrac12\norm{\bh_i}^2+y'a_i-\tfrac12a_i^2 ,
\]
obtained by cancelling the normalizing constants and expanding the two squares,
so that $q/\nu=\sum_i\lambda_ie^{f_i}$ and
$\psi_i=\lambda_ie^{f_i}/\sum_k\lambda_ke^{f_k}$. As in
\eqref{eq:fk}--\eqref{eq:gradpsi},
\begin{equation}\label{eq:onegrad}
\nabla_{\bx'}f_i=\bh_i+r_i\beta_i,\quad \partial_{y'}f_i=a_i,\quad
\nabla_{\bx'}\psi_i=\psi_i\big(\bh_i-\bar{\bh}+r_i\beta_i-\bar{\bm w}\big),\quad
\partial_{y'}\psi_i=\psi_i(a_i-\bar a).
\end{equation}

\emph{(a) Posterior identity.} For every $i$ with $\psi_i>0$ we have
$\log\frac q\nu=\log\frac{\lambda_i}{\psi_i}+f_i$. Averaging this identity with
the weights $\psi_i$ (which sum to one) and taking $\E_\nu$,
\begin{equation}\label{eq:postid}
\KL(\nu\|q)=-\E_\nu\log\frac q\nu
=\E_\nu\Big[\sum_i\psi_i\log\frac{\psi_i}{\lambda_i}\Big]-\E_\nu\Big[\sum_i\psi_if_i\Big].
\end{equation}

\emph{(b) Stein computations and an exact identity.} Two centring identities are
used repeatedly: since $\sum_i\psi_i(\bh_i-\bar{\bh})=0$ and
$\sum_i\psi_i(r_i\beta_i-\bar{\bm w})=0$,
\[
\sum_i\psi_i\ip{\bh_i-\bar{\bh},\bh_i}=V_h,\qquad
\sum_i\psi_i\ip{r_i\beta_i-\bar{\bm w},\bh_i}=\mathcal C,\qquad
\sum_i\psi_i(a_i-\bar a)a_i=V_y ,
\]
and likewise $\sum_i\psi_i\norm{\bh_i}^2=V_h+\norm{\bar{\bh}}^2$ and
$\sum_i\psi_ia_i^2=V_y+\bar a^2$. By Lemma~\ref{lem:stein} applied in $\bx'$ to
the field $F=\bar{\bh}=\sum_i\psi_i\bh_i$, and by \eqref{eq:onegrad},
\[
\E_\nu\ip{\bx',\bar{\bh}}=\E_\nu\sum_i\ip{\nabla_{\bx'}\psi_i,\bh_i}
=\E_\nu\sum_i\psi_i\ip{\bh_i-\bar{\bh}+r_i\beta_i-\bar{\bm w},\bh_i}=\E_\nu\big[V_h+\mathcal C\big],
\]
and by Lemma~\ref{lem:stein} in $y'$ with $\bx'$ fixed,
$\E_\nu\big[y'\sum_i\psi_ia_i\big]=\E_\nu\sum_i\partial_{y'}\psi_i\,a_i=\E_\nu V_y$.
Substituting these into $f_i$ and into \eqref{eq:Pone},
\begin{align}
\E_\nu\sum_i\psi_if_i&=\E_\nu\big[V_h+\mathcal C\big]-\tfrac12\E_\nu\big[V_h+\norm{\bar{\bh}}^2\big]+\E_\nu V_y-\tfrac12\E_\nu\big[V_y+\bar a^2\big]\notag\\
&=\E_\nu\Big[\tfrac12V_h+\mathcal C-\tfrac12\norm{\bar{\bh}}^2+\tfrac12V_y-\tfrac12\bar a^2\Big],\label{eq:Ef}\\
P&=\E_\nu\big[V_h+\norm{\bar{\bh}}^2\big]-\E_\nu\big[V_h+\mathcal C\big]+\E_\nu\big[V_y+\bar a^2\big]-\E_\nu V_y\notag\\
&=\E_\nu\big[\norm{\bar{\bh}}^2+\bar a^2-\mathcal C\big].\label{eq:Pid}
\end{align}
Inserting \eqref{eq:Ef} into \eqref{eq:postid}, multiplying by $2$ and using
\eqref{eq:Pid} to substitute $P$ for
$\E_\nu[\norm{\bar{\bh}}^2+\bar a^2-\mathcal C]$
gives the exact identity in \eqref{eq:exactP}.

\emph{(c) Log-Sobolev.} Gross's Gaussian logarithmic Sobolev inequality, in the
form
\[
\E_\nu\big[f^2\log f^2\big]-\E_\nu[f^2]\,\log\E_\nu[f^2]\ \le\ 2\,\E_\nu\norm{\nabla f}^2 ,
\]
becomes, after the substitution $F=f^2$ (so that
$\norm{\nabla f}^2=\norm{\nabla F}^2/(4F)$),
\[
\E_\nu[F\log F]-\E_\nu F\log\E_\nu F\ \le\ \tfrac12\,\E_\nu\big[\norm{\nabla F}^2/F\big]
\qquad\text{for positive smooth }F .
\]
Apply this to $F=\psi_i$, which is smooth and strictly positive, and for which
$\E_\nu\psi_i=\bar\lambda_i$, so that the left-hand side equals
$\E_\nu[\psi_i\log(\psi_i/\bar\lambda_i)]$. Summing over $i$ and using
\eqref{eq:onegrad},
\[
\sum_i\frac{\norm{\nabla\psi_i}^2}{\psi_i}
=\sum_i\psi_i\Big(\norm{(\bh_i-\bar{\bh})+(r_i\beta_i-\bar{\bm w})}^2+(a_i-\bar a)^2\Big)
=V_h+2\mathcal C+V_w+V_y ,
\]
which is the second part of \eqref{eq:exactP}. Since
$\sum_i\psi_i\log\frac{\psi_i}{\lambda_i}=\sum_i\psi_i\log\frac{\psi_i}{\bar\lambda_i}+\sum_i\psi_i\log\frac{\bar\lambda_i}{\lambda_i}$
and $\E_\nu\sum_i\psi_i\log\frac{\bar\lambda_i}{\lambda_i}=\KL(\bar\lambda\|\lambda)$,
combining the two parts of \eqref{eq:exactP} makes $V_h$ and $V_y$ cancel:
\[
P\ \ge\ 2\KL(\nu\|q)-2\KL(\bar\lambda\|\lambda)-\E_\nu\big[\mathcal C+V_w\big].
\]

\emph{(d) The deficit.} By Cauchy--Schwarz in the posterior and Young's
inequality $ab\le\frac\tau2a^2+\frac1{2\tau}b^2$,
\[
\abs{\mathcal C}\le\sum_i\psi_i\norm{r_i\beta_i-\bar{\bm w}}\,\norm{\bh_i-\bar{\bh}}
\le\frac\tau2V_h+\frac1{2\tau}V_w ,
\]
so $\E_\nu[\mathcal C+V_w]\le\frac\tau2\E_\nu V_h+(1+\frac1{2\tau})\E_\nu V_w$.
\end{proof}

\begin{lemma}[{Projection comparison and weight KL; formal version of Lemma~\ref{lem:sk-entropy}}]\label{lem:sk-entropy-app}
Under the assumptions of Proposition~\ref{prop:onestep-app}, for every
$\ell\in[m]$ and every $\tau\in(0,1]$,
\begin{gather*}
P_\ell\ \ge\ 2\KL(\vphi^*_\ell\|q_\ell)-2\KL(\bar\lambda^{(\ell)}\|\lambda^{(\ell)})-\tfrac{\tau}{2}B^2-6\Big(1+\tfrac1{2\tau}\Big)\Bth^2,\\[2pt]
\sum_\ell\pi^*_\ell\KL(\bar\lambda^{(\ell)}\|\lambda^{(\ell)})\le\varepsilon_\pi+\Ag e^{-c\Delta^2}.
\end{gather*}
\end{lemma}

\begin{proof}
\emph{First inequality.} Apply Lemma~\ref{lem:onecluster} to cluster $\ell$ in
the recentered coordinates; by Remark~\ref{rem:shear} its $P$ is $P_\ell$ and its
$\KL(\nu\|q)$ is $\KL(\vphi^*_\ell\|q_\ell)$. It remains to bound the two
variances. A variance is at most the corresponding second moment, so by
\eqref{eq:Bconseq}, $V_h\le\sum_i\psi_i\norm{\bh_i}^2\le B^2$. Similarly
$V_w\le\sum_i\psi_ir_i^2\norm{\beta_i}^2\le\Bth^2\sum_i\psi_ir_i^2$, and, using
$(u-v)^2\le2u^2+2v^2$ twice together with $\norm{\beta_i}\le\Bth$ and
$\abs{c_i}=\abs{\ip{\bmu^*_\ell,\beta_i}}\le\Dmax\Bth$,
\[
r_i^2\le2y'^2+2a_i^2\le2y'^2+4\Bth^2\big(\norm{\bx'}^2+\Dmax^2\big).
\]
Taking $\E_\nu$, so that $\E_\nu y'^2=1$ and $\E_\nu\norm{\bx'}^2=d$, and using
$\Bth^2\Lam\le1$ from Lemma~\ref{lem:constants}(ii),
\[
\E_\nu V_w\ \le\ \Bth^2\big(2+4\Bth^2\Lam\big)\ \le\ 6\Bth^2 .
\]

\emph{Second inequality.} Let $\bar q_\ell:=\sum_{i\in S_\ell}\bar\lambda^{(\ell)}_i\vphi_i$
and $\bar p:=\sum_\ell\hat\pi_\ell\bar q_\ell$, i.e.\ the model with weights
$\bar\pi_i:=\hat\pi_\ell\bar\lambda^{(\ell)}_i$ for $i\in S_\ell$; since
$\sum_{i\in S_\ell}\bar\lambda^{(\ell)}_i=1$, we have $\bar\bpi\in\cP$. By
Lemma~\ref{lem:dualgap},
$\E_{p_*}\log\frac{\bar p}{p}=\cL(\bpi,\bmu,\btheta)-\cL(\bar\bpi,\bmu,\btheta)\le\varepsilon_\pi$.
Conversely, for $\bz$ from cluster $\ell$ we have $\bar p\ge\hat\pi_\ell\bar q_\ell$
and $p=\hat\pi_\ell q_\ell(1+\chi_\ell)$, so, using
$\vphi_i/q_\ell=\psi^{(\ell)}_i/\lambda^{(\ell)}_i$ and then Jensen's inequality
for the concave logarithm against the probability vector $\psi^{(\ell)}$,
\[
\log\frac{\bar p}{p}\ \ge\ \log\frac{\bar q_\ell}{q_\ell}-\log(1+\chi_\ell)
=\log\sum_{i\in S_\ell}\psi^{(\ell)}_i\frac{\bar\lambda^{(\ell)}_i}{\lambda^{(\ell)}_i}-\log(1+\chi_\ell)
\ \ge\ \sum_{i\in S_\ell}\psi^{(\ell)}_i\log\frac{\bar\lambda^{(\ell)}_i}{\lambda^{(\ell)}_i}-\log(1+\chi_\ell).
\]
Taking $\E_\ell$ turns $\psi^{(\ell)}_i$ into $\bar\lambda^{(\ell)}_i$ and hence
the first sum into $\KL(\bar\lambda^{(\ell)}\|\lambda^{(\ell)})$; multiplying by
$\pi^*_\ell$, summing over $\ell$ and using Lemma~\ref{lem:cross}(3) for the
interference term gives
$\varepsilon_\pi\ge\sum_\ell\pi^*_\ell\KL(\bar\lambda^{(\ell)}\|\lambda^{(\ell)})-\Ag e^{-c\Delta^2}$.
\end{proof}

\subsection{Loss versus within-group KL divergence}\label{sec:applossKL}

\begin{lemma}[Loss vs.\ within-group KL; formal version of Lemma~\ref{lem:sk-lossKL}]\label{lem:lossKL-app}
Under the assumptions of Proposition~\ref{prop:onestep-app},
\[
\cL(\bpi,\bmu,\btheta)\ \le\ \sum_\ell\pi^*_\ell\KL(\vphi^*_\ell\|q_\ell)+C\varepsilon_\pi+\Ag e^{-c\Delta^2}.
\]
\end{lemma}

\begin{proof}
For $\bz$ from cluster $\ell$, factoring the local component out of each mixture
gives $p_*=\pi^*_\ell\vphi^*_\ell(1+\chi^*_\ell)$, while dropping all but the
$\ell$th group of the model gives $p\ge\hat\pi_\ell q_\ell$. Hence
\[
\log\frac{p_*}{p}\ \le\ \log\frac{\vphi^*_\ell}{q_\ell}+\log\frac{\pi^*_\ell}{\hat\pi_\ell}+\log(1+\chi^*_\ell).
\]
Taking $\E_\ell$, multiplying by $\pi^*_\ell$ and summing over $\ell$, and noting
that the middle term is deterministic,
\[
\cL\ \le\ \sum_\ell\pi^*_\ell\KL(\vphi^*_\ell\|q_\ell)+\sum_\ell\pi^*_\ell\log\frac{\pi^*_\ell}{\hat\pi_\ell}+\sum_\ell\pi^*_\ell\E_\ell\log(1+\chi^*_\ell).
\]
By Lemma~\ref{lem:groupweight}, $\hat\pi_\ell\ge\pi^*_\ell/2$, so $\log u\le u-1$
with $u=\pi^*_\ell/\hat\pi_\ell$ gives
$\log\frac{\pi^*_\ell}{\hat\pi_\ell}\le\frac{\pi^*_\ell-\hat\pi_\ell}{\hat\pi_\ell}\le\frac{2\abs{\pi^*_\ell-\hat\pi_\ell}}{\pi^*_\ell}$;
the factors $\pi^*_\ell$ cancel and the middle sum is at most
$2\sum_\ell\abs{\hat\pi_\ell-\pi^*_\ell}\le4\varepsilon_\pi+\Ag e^{-c\Delta^2}$.
The last sum is at most $\Ag e^{-c\Delta^2}$ by Lemma~\ref{lem:cross}(4), since
$\sum_\ell\pi^*_\ell=1$.
\end{proof}
\section{Regularity and Gaussian Ratio Estimates}\label{sec:appD}

\subsection{Local smoothness and self-bounding of the gradient}\label{sec:appsmooth}

\begin{lemma}[Local smoothness]\label{lem:smooth}
Fix $\bpi$ and $(\bmu,\btheta)$ and suppose that, for every $\ell\in[m]$,
$\hat\pi_\ell\ge c_{\mathrm{cov}}\pi^*_\ell$ and
\[
\frac1{\hat\pi_\ell}\sum_{i\in S_\ell}\pi_i\big(\norm{\bmu_i-\bmu^*_\ell}^2+\norm{\btheta_i-\btheta^*_\ell}^2\big)\le B^2,
\qquad \norm{\btheta_i-\btheta^*_\ell}\le1\ \ (i\in S_\ell).
\]
Let $\bm\delta=(\bm\delta^\mu_i,\bm\delta^\theta_i)_{i\le n}$ satisfy
$\frac1{\hat\pi_\ell}\sum_{i\in S_\ell}\pi_i\norm{\bm\delta_i}^2\le D^2$ for every
$\ell$, and set
\begin{gather*}
A(D):=C(1+\Lam)^2\Big(1+d+B^2+D^2+\log\tfrac1{c_{\mathrm{cov}}\pimin}\Big),\\[2pt]
A:=C(1+\Lam)^2\Big(1+d+B^2+\log\tfrac1{c_{\mathrm{cov}}\pimin}+\tfrac1{c_{\mathrm{cov}}\pimin}\Big)
\end{gather*}
with $C$ sufficiently large; for $c_{\mathrm{cov}}=\tfrac12$ this $A$ is the
constant \eqref{eq:Aconst}. Then, writing
$g:=\nabla_{\bmu,\btheta}\cL(\bpi,\bmu,\btheta)$,
\begin{align}
\cL(\bpi,(\bmu,\btheta)+\bm\delta)&\le\cL(\bpi,\bmu,\btheta)+\ip{g,\bm\delta}+\tfrac12A(D)\norm{\bm\delta}_F^2,\label{eq:taylor}\\
\norm{g}_F^2&\le A,\label{eq:gbound}\\
\cL(\bpi,(\bmu,\btheta)-\eta g)&\le\cL(\bpi,\bmu,\btheta)-\tfrac\eta2\norm{g}_F^2\qquad(\eta\le c/A),\label{eq:descentineq}\\
\norm{g}_F^2&\le2A\,\cL(\bpi,\bmu,\btheta).\label{eq:selfbound}
\end{align}
\end{lemma}

\begin{proof}
\emph{Hessian bound.} For a single sample $\bz$ let
$f_{\bz}(\bmu,\btheta):=-\log\sum_i\pi_i\vphi_i(\bz)$, so that
$\cL=\E_{p_*}f_{\bz}+\text{const}$. Fix a direction $\bm v$, put
$F(t):=f_{\bz}\big((\bmu,\btheta)+t\bm v\big)$ and
$\ell_i(t):=\log\pi_i+\log\vphi_i(\bz)$ evaluated at $(\bmu,\btheta)+t\bm v$.
Differentiating
$\log\vphi_i=\text{const}-\tfrac12\norm{\bx-\bmu_i}^2-\tfrac12r_i^2$ along $\bm v$,
\[
\ell_i'=\ip{\bx-\bmu_i(t),\bm v^\mu_i}+r_i(t)\ip{\bx,\bm v^\theta_i},
\qquad
\ell_i''=-\norm{\bm v^\mu_i}^2-\ip{\bx,\bm v^\theta_i}^2 .
\]
Since $F=-\log\sum_ie^{\ell_i}$, we have $F'=-\sum_i\psi_i\ell_i'$ and, using
$\frac{d}{dt}\psi_i=\psi_i(\ell_i'-\sum_k\psi_k\ell_k')$,
\[
F''=-\sum_i\psi_i\ell_i''-\Big[\sum_i\psi_i(\ell_i')^2-\Big(\sum_i\psi_i\ell_i'\Big)^{\!2}\Big].
\]
The bracket is the $\psi$-variance of $\ell'$, hence nonnegative, and
$\ell_i''\le0$, so
$-\sum_i\psi_i(\ell_i')^2\le F''\le\sum_i\psi_i\big(\norm{\bm v^\mu_i}^2+\norm{\bx}^2\norm{\bm v^\theta_i}^2\big)$.
With $(\ell_i')^2\le2\big(\norm{\bx-\bmu_i(t)}^2+r_i(t)^2\norm{\bx}^2\big)\norm{\bm v_i}^2$
by $(a+b)^2\le2a^2+2b^2$ and Cauchy--Schwarz, taking the supremum over
$\norm{\bm v}_F\le1$ gives
\begin{equation}\label{eq:hessop}
\begin{aligned}
\norm{\nabla^2f_{\bz}}_{\op}
&\le1+\norm{\bx}^2+2\sum_i\psi_i\big(\norm{\bx-\bmu_i(t)}^2+r_i(t)^2\norm{\bx}^2\big)\\
&\le(1+\norm{\bx}^2)\Big(1+2\sum_i\psi_i\big(\norm{\bx-\bmu_i(t)}^2+r_i(t)^2\big)\Big),
\end{aligned}
\end{equation}
all posteriors being evaluated at $(\bmu,\btheta)+t\bm v$; the last step uses
$a+b\norm{\bx}^2\le(a+b)(1+\norm{\bx}^2)$ for $a,b\ge0$.

To control the remaining sum we use the Gibbs variational principle
$-\log\sum_i\pi_ie^{-\alpha_i}=\min_{\bm\varrho\in\cP}\big\{\sum_i\varrho_i\alpha_i+\sum_i\varrho_i\log(\varrho_i/\pi_i)\big\}$
with $\alpha_i:=\tfrac12\norm{\bx-\bmu_i(t)}^2+\tfrac12r_i(t)^2$, whose minimizer
is $\varrho=\psi$. Evaluating the objective at the feasible point
$\varrho_i=\pi_i/\hat\pi_\ell$ on $S_\ell$ and $\varrho_i=0$ elsewhere, whose
entropy term is exactly $\log(1/\hat\pi_\ell)$, and dropping the nonnegative term
$\sum_i\psi_i\log(\psi_i/\pi_i)$ on the left,
\begin{equation}\label{eq:gibbs}
\sum_i\psi_i\big(\norm{\bx-\bmu_i(t)}^2+r_i(t)^2\big)
\ \le\ \frac1{\hat\pi_\ell}\sum_{i\in S_\ell}\pi_i\big(\norm{\bx-\bmu_i(t)}^2+r_i(t)^2\big)+2\log\frac1{\hat\pi_\ell}.
\end{equation}

Now take $\bz$ from cluster $\ell$, $\bm v=\bm\delta$ and $t\in[0,1]$, and write
$\bh_i(t):=\bmu_i(t)-\bmu^*_\ell$, $\beta_i(t):=\btheta_i(t)-\btheta^*_\ell$, so
that $\norm{\bh_i(t)}^2\le2\norm{\bh_i}^2+2\norm{\bm\delta^\mu_i}^2$ and
$\norm{\beta_i(t)}^2\le2\norm{\beta_i}^2+2\norm{\bm\delta^\theta_i}^2$. Apply
Cauchy--Schwarz to \eqref{eq:hessop} in the form
$\E_\ell[XY]\le\norm{X}_{L^2}\norm{Y}_{L^2}$ with $X=1+\norm{\bx}^2$ and $Y$ the
second factor. By Lemma~\ref{lem:moments}, $\norm{X}_{L^2}\le C(1+\Lam)$. For
$Y$, insert \eqref{eq:gibbs} and apply Minkowski's inequality, which moves the
$L^2$ norm inside the convex combination:
\[
\norm{Y}_{L^2}\ \le\ 1+4\log\frac1{\hat\pi_\ell}+\frac{2}{\hat\pi_\ell}\sum_{i\in S_\ell}\pi_i\big\|\,\norm{\bx-\bmu_i(t)}^2+r_i(t)^2\big\|_{L^2}.
\]
Since $\E_\ell\norm{\bx-\bmu_i(t)}^4\le C\big(d^2+\norm{\bh_i(t)}^4\big)$ and,
because $r_i(t)=\epsilon-\bx^\top\beta_i(t)$ with
$\bx^\top\beta_i(t)\sim N(\ip{\bmu^*_\ell,\beta_i(t)},\norm{\beta_i(t)}^2)$,
$\E_\ell r_i(t)^4\le C\big(1+\Lam^2\norm{\beta_i(t)}^4\big)$ by
Lemma~\ref{lem:moments}, subadditivity of the square root gives
$\big\|\norm{\bx-\bmu_i(t)}^2+r_i(t)^2\big\|_{L^2}\le C\big(1+d+\norm{\bh_i(t)}^2+\Lam\norm{\beta_i(t)}^2\big)$.
Averaging over $i\in S_\ell$ with weights $\pi_i/\hat\pi_\ell$, using $\Lam\ge1$
and the two hypotheses on $B^2$ and $D^2$,
\[
\frac1{\hat\pi_\ell}\sum_{i\in S_\ell}\pi_i\big(1+d+\norm{\bh_i(t)}^2+\Lam\norm{\beta_i(t)}^2\big)
\ \le\ 1+d+2\Lam\big(B^2+D^2\big)\ \le\ 2\Lam\big(1+d+B^2+D^2\big).
\]
Finally $\log(1/\hat\pi_\ell)\le\log\big(1/(c_{\mathrm{cov}}\pimin)\big)$, so
$\norm{Y}_{L^2}\le C(1+\Lam)\big(1+d+B^2+D^2+\log\frac1{c_{\mathrm{cov}}\pimin}\big)$
and therefore $\E_\ell\norm{\nabla^2f_{\bz}}_{\op}\le A(D)$. Averaging over $\ell$
with weights $\pi^*_\ell$ gives $\norm{\nabla^2\cL}_{\op}\le A(D)$ along the whole
segment (the groups $S_\ell$ are those of the base point and are used only to
choose the comparison $\varrho$; no reassignment is needed). Taylor's theorem
with integral remainder,
$\cL(1)=\cL(0)+\ip{g,\bm\delta}+\int_0^1(1-t)\,\frac{d^2}{dt^2}\cL\,dt$, and
$\int_0^1(1-t)dt=\tfrac12$ give \eqref{eq:taylor}.

\emph{Gradient bound.} By \eqref{eq:grad}, Jensen's inequality
$\norm{\E X}^2\le\E\norm{X}^2$ and $\psi_i^2\le\psi_i$,
\[
\norm{g}_F^2=\sum_i\Big(\big\|\E_{p_*}[\psi_i(\bmu_i-\bx)]\big\|^2+\big\|\E_{p_*}[\psi_ir_i\bx]\big\|^2\Big)
\le\E_{p_*}\Big[\sum_i\psi_i\big(\norm{\bx-\bmu_i}^2+r_i^2\norm{\bx}^2\big)\Big]\le A,
\]
the last step being the estimate just performed with $t=0$ and $\bm\delta=0$,
which yields $A(0)\le A$.

\emph{Descent and self-bound.} For $\bm\delta=-\eta g$ with $\eta\le c/A$,
\[
\frac1{\hat\pi_\ell}\sum_{i\in S_\ell}\pi_i\norm{\bm\delta_i}^2
\le\frac{\eta^2\norm{g}_F^2}{c_{\mathrm{cov}}\pimin}\le\frac{c^2}{A\,c_{\mathrm{cov}}\pimin}\le1,
\]
because $A$ contains the term $1/(c_{\mathrm{cov}}\pimin)$ with a large constant.
Hence \eqref{eq:taylor} applies with $D=1$, and $A(1)\le A$ by the same
observation, giving
$\cL((\bmu,\btheta)-\eta g)\le\cL-\eta\norm{g}_F^2+\frac{A\eta^2}2\norm{g}_F^2\le\cL-\frac\eta2\norm{g}_F^2$
for $\eta\le1/A$, which is \eqref{eq:descentineq}. Taking $\eta=1/A$ and using
$\cL\ge0$ on the left-hand side gives $0\le\cL-\frac1{2A}\norm{g}_F^2$, i.e.\
\eqref{eq:selfbound}.
\end{proof}

\subsection{Gaussian products and the regression factor}\label{sec:appgauss}

\begin{lemma}[Gaussian products]\label{lem:gaussprod}
Let $p\ge1$, let $s_1,\dots,s_K\in\R$ with $\sum_ks_k=1$, and let
$\bm m_1,\dots,\bm m_K\in\R^p$. Then, for all $\bx\in\R^p$,
\[
\prod_k\vphi(\bx;\bm m_k)^{s_k}
=\exp\Big(-\tfrac12\sum_{k<l}s_ks_l\norm{\bm m_k-\bm m_l}^2\Big)\,\vphi\Big(\bx;\bar{\bm m}\Big),
\qquad \bar{\bm m}:=\sum_ks_k\bm m_k .
\]
Consequently, for any weight $W$,
\[
\int\prod_k\vphi(\bx;\bm m_k)^{s_k}W(\bx)\,d\bx
=\exp\Big(-\tfrac12\sum_{k<l}s_ks_l\norm{\bm m_k-\bm m_l}^2\Big)\,\E_{\bx\sim N(\bar{\bm m},I_p)}W(\bx).
\]
\end{lemma}

\begin{proof}
Using $\sum_ks_k=1$ and completing the square,
\[
\sum_ks_k\norm{\bx-\bm m_k}^2
=\norm{\bx}^2-2\ip{\bx,\bar{\bm m}}+\sum_ks_k\norm{\bm m_k}^2
=\norm{\bx-\bar{\bm m}}^2+\Big(\sum_ks_k\norm{\bm m_k}^2-\norm{\bar{\bm m}}^2\Big).
\]
Expanding the double sum and using $\sum_ks_k=1$ twice,
\[
\tfrac12\sum_{k,l}s_ks_l\norm{\bm m_k-\bm m_l}^2
=\tfrac12\sum_ks_k\norm{\bm m_k}^2+\tfrac12\sum_ls_l\norm{\bm m_l}^2-\ip{\bar{\bm m},\bar{\bm m}}
=\sum_ks_k\norm{\bm m_k}^2-\norm{\bar{\bm m}}^2 ,
\]
and the diagonal terms vanish while off-diagonal pairs are counted twice, so
$\tfrac12\sum_{k,l}=\sum_{k<l}$. The normalizing constants multiply to
$\big((2\pi)^{-p/2}\big)^{\sum_ks_k}=(2\pi)^{-p/2}$, which together with
$e^{-\norm{\bx-\bar{\bm m}}^2/2}$ reconstitutes $\vphi(\bx;\bar{\bm m})$. The
second statement follows by pulling the constant out of the integral and
recognizing the remaining integral as an expectation.
\end{proof}

Applied in the $y$ variable ($p=1$), Lemma~\ref{lem:gaussprod} is the only place
where the regression factor enters the cross-group analysis. For fixed $\bx$ we
write $u_\bullet:=\bx^\top\btheta_\bullet$ for the conditional mean of $y$ under
the component $\bullet$.

\begin{lemma}[Regression factors]\label{lem:yratio}
Let $j$ be the data cluster, $\ell\ne j$, $i\in S_\ell$, $k\in S_j$ and $r\ne j$,
and assume \eqref{eq:Bconseq}. For fixed $\bx$, with $\kappa$ as in
\eqref{eq:kappa}:
\begin{enumerate}[nosep]
\item $\displaystyle\int\vphi_1(y;u^*_j)\,\vphi_1(y;u_i)^{1/2}\,\vphi_1(y;u_k)^{-1/2}\,dy\le e^{\kappa\norm{\bx}^2/2}$,
      and the Gaussian produced by Lemma~\ref{lem:gaussprod} is centered at
      $\bar u$ with $\abs{\bar u-u^*_j}\le(1+\Dth)\norm{\bx}$;
\item $\displaystyle\int\vphi_1(y;u^*_j)^{1/2}\,\vphi_1(y;u_i)\,\vphi_1(y;u^*_\ell)^{-1/2}\,dy\le e^{\kappa\norm{\bx}^2/2}$;
\item $\displaystyle\int\vphi_1(y;u^*_j)^{1/2}\vphi_1(y;u^*_r)^{1/2}\,dy=e^{-(u^*_j-u^*_r)^2/8}\le1$.
\end{enumerate}
\end{lemma}

\begin{proof}
All three are instances of Lemma~\ref{lem:gaussprod} with $p=1$ and $W$ being constant;
the resulting Gaussian integrates to one, so each integral equals the exponential
prefactor $\exp\big(-\tfrac12\sum_{k<l}s_ks_l(u_k-u_l)^2\big)$.

(1) Here $s=(1,\tfrac12,-\tfrac12)$, which sums to one, and the exponent is
$-\tfrac12\big[\tfrac12(u^*_j-u_i)^2-\tfrac12(u^*_j-u_k)^2-\tfrac14(u_i-u_k)^2\big]$.
Put $a:=u^*_j-u_k$ and $b:=u^*_j-u_i$, so $u_i-u_k=a-b$; the exponent equals
\[
-\tfrac14b^2+\tfrac14a^2+\tfrac18(a-b)^2=\tfrac38a^2-\tfrac14ab-\tfrac18b^2\ \le\ \abs a\big(\abs a+\abs b\big).
\]
By \eqref{eq:Bconseq} and Cauchy--Schwarz, $\abs a\le\Bth\norm{\bx}$ (as
$k\in S_j$) and $\abs b\le(\Dth+\Bth)\norm{\bx}$ (as $i\in S_\ell$, via
$\norm{\btheta^*_j-\btheta_i}\le\Dth+\Bth$), whence
$\abs a(\abs a+\abs b)\le\Bth(\Dth+2\Bth)\norm{\bx}^2=\kappa\norm{\bx}^2/2$. The
center is $\bar u=u^*_j+\tfrac12(u_i-u_k)$ with
$\abs{u_i-u_k}\le(\Dth+2\Bth)\norm{\bx}$, so
$\abs{\bar u-u^*_j}\le\tfrac12(\Dth+2)\norm{\bx}\le(1+\Dth)\norm{\bx}$ using
$\Bth\le1$.

(2) Here $s=(\tfrac12,1,-\tfrac12)$. With $e:=u^*_j-u^*_\ell$ and
$\delta:=u_i-u^*_\ell$, so that $\abs\delta\le\Bth\norm{\bx}$ and
$\abs e\le\Dth\norm{\bx}$, the exponent is
$-\tfrac14(e-\delta)^2+\tfrac18e^2+\tfrac14\delta^2=-\tfrac18e^2+\tfrac12e\delta\le\tfrac12\Dth\Bth\norm{\bx}^2\le\kappa\norm{\bx}^2/2$.

(3) Here $s=(\tfrac12,\tfrac12)$ and the prefactor is the coefficient $e^{-(u^*_j-u^*_r)^2/8}\le1$.
\end{proof}

\subsection{The cell ratio bounds}\label{sec:appratio}

\begin{lemma}[Gaussian ratio bounds]\label{lem:ratio}
Assume \eqref{eq:Bconseq}, $\kappa(\Lam+B^2)\le c_0$ with $c_0$ small, and
$\Delta\ge32B$. Fix a degree $K$. Then for every data cluster $j$, every
$\ell\ne j$, $i\in S_\ell$, $k\in S_j$, $r\ne j$, and every
$W_K(\bz)=(1+\norm{\bx-\bmu^*_j}+\abs{\epsilon})^K$ with
$\epsilon=y-\bx^\top\btheta^*_j$, the three bounds \eqref{eq:ratiorecap} hold with
$\Ag=C_K\big((1+B)(1+\Dmax)(1+\Dth)(1+d)\big)^{CK}$ and $c=1/16$.
\end{lemma}

\begin{proof}
Write $D:=\norm{\bmu^*_\ell-\bmu^*_j}\ge\Delta$ and $\bm Z:=\bx-\bmu^*_j$; the
hypotheses give $D\ge\Delta\ge32B$ and, by Lemma~\ref{lem:constants}(iii),
$2\kappa\le\tfrac12$.

\emph{First bound.} Integrate $y$ first. By Lemma~\ref{lem:yratio}(1) and
Lemma~\ref{lem:gaussprod}, for fixed $\bx$,
\begin{align*}
\int\vphi_1(y;u^*_j)\vphi_1(y;u_i)^{1/2}&\vphi_1(y;u_k)^{-1/2}\big(1+\norm{\bm Z}+\abs{y-u^*_j}\big)^K dy\\
&\le\ e^{\kappa\norm{\bx}^2/2}\;\E_{y\sim N(\bar u,1)}\big(1+\norm{\bm Z}+\abs{y-u^*_j}\big)^K .
\end{align*}
Inside the expectation, $\abs{y-u^*_j}\le\abs{y-\bar u}+(1+\Dth)\norm{\bx}$ by
Lemma~\ref{lem:yratio}(1), and the elementary inequality $A+B\le A(1+B)$ for
$A\ge1$, $B\ge0$, applied twice, gives
\[
1+\norm{\bm Z}+\abs{y-u^*_j}\ \le\ (1+\norm{\bm Z})\big(1+(1+\Dth)\norm{\bx}\big)\big(1+\abs{y-\bar u}\big).
\]
Raising to the $K$th power, factoring out the two deterministic factors, and
using that $y-\bar u$ is standard normal, so that
$\E(1+\abs{y-\bar u})^K=:C_K$ is an absolute constant, the $y$ integral is at
most $e^{\kappa\norm{\bx}^2/2}W'(\bx)$ with
\[
W'(\bx):=C_K(1+\norm{\bm Z})^K\big(1+(1+\Dth)\norm{\bx}\big)^K .
\]

It remains to bound
$\int\vphi(\bx;\bmu^*_j)\vphi(\bx;\bmu_i)^{1/2}\vphi(\bx;\bmu_k)^{-1/2}e^{\kappa\norm{\bx}^2/2}W'(\bx)\,d\bx$.
By Lemma~\ref{lem:gaussprod} with $s=(1,\tfrac12,-\tfrac12)$ this equals
$M\cdot\E_{\bx\sim N(\bm c,I)}\big[e^{\kappa\norm{\bx}^2/2}W'(\bx)\big]$ with
$\bm c=\bmu^*_j+\tfrac12(\bmu_i-\bmu_k)$ and
\begin{align*}
M&=\exp\Big(-\tfrac14\norm{\bmu^*_j-\bmu_i}^2+\tfrac14\norm{\bmu^*_j-\bmu_k}^2+\tfrac18\norm{\bmu_i-\bmu_k}^2\Big)\\
&\le\exp\Big(-\tfrac14(D-B)^2+\tfrac14B^2+\tfrac18(D+2B)^2\Big)=e^{-D^2/8+DB+B^2/2},
\end{align*}
using $\norm{\bmu^*_j-\bmu_i}\ge D-B$, $\norm{\bmu^*_j-\bmu_k}\le B$ and
$\norm{\bmu_i-\bmu_k}\le D+2B$, all consequences of \eqref{eq:Bconseq} and the
triangle inequality.

For the expectation, $\norm{\bm c}\le2\Dmax+B$ (since
$\norm{\bmu_i},\norm{\bmu_k}\le\Dmax+B$), and $\norm{\bx}^2$ is noncentral
chi-square under $N(\bm c,I)$, so for $2\kappa\le\tfrac12$,
\[
\E e^{\kappa\norm{\bx}^2}=(1-2\kappa)^{-d/2}\exp\Big(\frac{\kappa\norm{\bm c}^2}{1-2\kappa}\Big)
\le e^{2\kappa d+2\kappa(2\Dmax+B)^2}\le e^{C\kappa(\Lam+B^2)}\le e^{Cc_0}\le C ,
\]
where $-\log(1-u)\le2u$ for $u\le\tfrac12$ was used. Moreover, since
$\norm{\bx-\bmu^*_j}\le\norm{\bm c-\bmu^*_j}+\norm{\bx-\bm c}\le\Dmax+B+\norm{\bx-\bm c}$
and $\norm{\bx}\le2\Dmax+B+\norm{\bx-\bm c}$, factoring as above and using that
$\norm{\bx-\bm c}$ has moments of every order $p\le4K$ bounded by
$C_K(1+d)^{2K}$ gives
$\E W'(\bx)^2\le C_K\big((1+\Dth)(1+\Dmax)(1+B)(1+d)\big)^{CK}$. Cauchy--Schwarz
therefore yields $\E\big[e^{\kappa\norm{\bx}^2/2}W'\big]\le\Ag$, so the first
bound is at most $\Ag e^{-D^2/8+DB+B^2/2}\le\Ag e^{-D^2/16}$: indeed, for
$B\le D/32$,
\[
-\tfrac18D^2+DB+\tfrac12B^2\ \le\ -\tfrac1{16}D^2 .
\]

\emph{Second bound.} Identical, with Lemma~\ref{lem:yratio}(2) for the $y$
integral and Lemma~\ref{lem:gaussprod} with $s=(\tfrac12,1,-\tfrac12)$ for the
$\bx$ integral, whose constant is
\begin{align*}
M&=\exp\Big(-\tfrac14\norm{\bmu^*_j-\bmu_i}^2+\tfrac18D^2+\tfrac14\norm{\bmu_i-\bmu^*_\ell}^2\Big)\\
&\le\exp\Big(-\tfrac14(D-B)^2+\tfrac18D^2+\tfrac14B^2\Big)=e^{-D^2/8+DB/2},
\end{align*}
and whose center $\bm c=\tfrac12\bmu^*_j+\bmu_i-\tfrac12\bmu^*_\ell$ again
satisfies $\norm{\bm c}\le2\Dmax+B$; the factor
$\E e^{\kappa\norm{\bx}^2/2}\le C$ is bounded as before.

\emph{Third bound.} As a density in $\bz$, $\vphi^*_r$ is Gaussian with a covariance that depends on $\btheta^*_r$, so Lemma~\ref{lem:gaussprod} does not apply in $\bz$ directly. Integrating $y$ first, Lemma~\ref{lem:yratio}(3) bounds the inner integral by one, and Lemma~\ref{lem:gaussprod} in $\bx$ then gives
$\E_j\big[(\vphi^*_r/\vphi^*_j)^{1/2}\big]=\int\sqrt{\vphi^*_r\vphi^*_j}\,d\bz\le\int\sqrt{\vphi(\bx;\bmu^*_r)\vphi(\bx;\bmu^*_j)}\,d\bx=e^{-\norm{\bmu^*_r-\bmu^*_j}^2/8}\le e^{-\Delta^2/8}$.
\end{proof}
\section{Additional Results}\label{sec:appE}

\subsection{The expert part of the projection is a quadratic form}

\begin{proposition}\label{prop:quadform}
Fix $\ell$ and write
$\bar\beta^{(\ell)}(\bz):=\sum_{i\in S_\ell}\psi^{(\ell)}_i(\bz)(\btheta_i-\btheta^*_\ell)$.
Then the expert part of the one-cluster projection \eqref{eq:Pell} satisfies
\[
P_{\ell,\theta}
=-\E_\ell\Big[\sum_{i\in S_\ell}\psi^{(\ell)}_i(\bz)\,r_i(\bz)\,\bx^\top(\btheta_i-\btheta^*_\ell)\Big]
=\E_\ell\Big[\big(\bx^\top\bar\beta^{(\ell)}(\bz)\big)^2\Big]\ \ge\ 0 .
\]
In particular, up to the cross-group errors of Lemma~\ref{lem:decouple}, a
gradient step never moves the experts of a group away from the true expert of
that group on average.
\end{proposition}

\begin{proof}
Work in the recentered coordinates of Remark~\ref{rem:shear}. By
\eqref{eq:Tirecenterd}, $P_{\ell,\theta}=\E_\nu\big[\sum_i\psi_i(a_i^2-y'a_i)\big]$
with $a_i=\bx^\top(\btheta_i-\btheta^*_\ell)$. The Stein computation in part~(b)
of the proof of Lemma~\ref{lem:onecluster} gives
$\E_\nu\big[y'\sum_i\psi_ia_i\big]=\E_\nu V_y$, while
$\sum_i\psi_ia_i^2=V_y+\bar a^2$. The variances cancel and
$P_{\ell,\theta}=\E_\nu\bar a^2=\E_\ell\big[(\bx^\top\bar\beta^{(\ell)})^2\big]\ge0$.
\end{proof}

\subsection{Invariance of the groups during the global phase}

\begin{proposition}\label{prop:invariance}
Under the assumptions of Proposition~\ref{prop:onestep-app} and $\eta\le c/A$,
every $i\in S_\ell$ satisfies $\norm{\bmu_i-\bmu^*_\ell}\le B<\Delta/4$ and
$\norm{\bmu^+_i-\bmu^*_\ell}\le B+1<\Delta/2$; consequently
$S_\ell(\bmu^+)=S_\ell(\bmu)$ for every $\ell$.
\end{proposition}

\begin{proof}
By \eqref{eq:Bconseq}, $\norm{\bmu_i-\bmu^*_\ell}\le B$ for $i\in S_\ell$, and
$B<\Delta/4$ by Lemma~\ref{lem:constants}(iv). The single-component gradient is a
block of the full gradient, so
$\norm{\nabla_{\bmu_i}\cL}\le\norm{g}_F\le\sqrt{2AL_0}$ by \eqref{eq:selfbound}
and $\cL\le L_0$. Hence, by the triangle inequality and $\eta\le c/A$,
\[
\norm{\bmu^+_i-\bmu^*_\ell}\ \le\ \norm{\bmu_i-\bmu^*_\ell}+\eta\norm{\nabla_{\bmu_i}\cL}
\ \le\ B+\frac{c}{A}\sqrt{2AL_0}=B+c\sqrt{\frac{2L_0}{A}}\ \le\ B+1 ,
\]
because $A\ge L_0$ by \eqref{eq:AgeL} and $c$ is small. For $j\ne\ell$, the
reverse triangle inequality gives
$\norm{\bmu^+_i-\bmu^*_j}\ge\Delta-(B+1)>B+1\ge\norm{\bmu^+_i-\bmu^*_\ell}$,
where the strict inequality is $\Delta>2(B+1)$ from
Lemma~\ref{lem:constants}(iv). Thus $\bmu^*_\ell$ remains the strictly nearest
true center of $\bmu^+_i$, and no component changes group.
\end{proof}

\section{Experimental details}\label{sec:expdetails}

All experiments sample data from \eqref{eq:truth} with $d=10$ and run Algorithm~\ref{alg:main} with the population expectations replaced by averages over $N=10{,}000$ training points, for at most $T=10{,}000$ iterations. Table~\ref{tab:hparams} lists the settings of each experiment.

\emph{Mixing-weight update.} Each outer iteration performs four EM fixed-point updates of the fitted weights, holding the centers and experts fixed: $\pi_i\leftarrow N^{-1}\sum_{k=1}^N\psi_i(\bz_k)$, where $\psi_i$ is the current joint responsibility. These four updates are not an exact solve of the weight subproblem. The balancing and pruning runs penalize $\lambda\KL(\bar{\boldsymbol\rho}\|\mathrm{Unif}_n)$, where $\bar\rho_i=N^{-1}\sum_k P(i\mid\bx_k)$ is the mean input-only router probability on training data. Their four inner weight updates use safeguarded generalized EM: an ordinary EM proposal and a backtracked gradient proposal are evaluated against the penalized objective, and an improving proposal is retained. The penalty used here is distinct from $\lambda\KL(\mathrm{Unif}_n\|\bpi)$; the latter prior-weight penalty was not used in these runs. Centers and experts take one gradient step per outer iteration. The heatmap uses a fixed step size; the other experiments start each step at $0.02$ and backtrack as needed.

\emph{Ground truth.} In the heatmap, centers are proposed sequentially from $\mathcal N(0,(8/\sqrt{2})^2I_d)$ and accepted only if they are at least $6.4$ from every previously accepted center; the accepted centers are then recentered. Thus $8$ is the placement scale, while the realized minimum separation is at least $6.4$, not exactly $8$. This rejection construction also applies when $m=50$. Heatmap experts have independent $\mathcal N(0,0.5^2I_d)$ coefficients, and its true mixing weights follow $\operatorname{Dirichlet}(3,\ldots,3)$.

For the other experiments, the centers are a randomly rotated regular $m$-gon, with independent Gaussian coordinate jitter of standard deviation $0.08$, recentered and rescaled so that their minimum pairwise distance equals $\Delta$ exactly. True expert coefficients are independent standard Gaussians, centered across experts and rescaled to coordinate root-mean-square $s_\theta$, as specified in Table~\ref{tab:hparams}. The warm-start experiment uses uniform true mixing weights. The balancing and pruning experiments use either uniform weights or the skewed prior
\[
\bpi^*=(0.000001,\,0.07367372,\,0.11166848,\,0.16925777,\,0.25654680,\,0.38885223),
\]
with $\sum_\ell(\pi^*_\ell)^2=0.2636$, i.e.\ $n^*_{\mathrm{eff}}=3.794$. Its smallest component has expected training count $0.01$; the skewed setting does not measure reliable recovery of that rare expert. The local least-squares start fits $\btheta_i^{(0)}$ on the observations within radius $R$ of $\bmu_i^{(0)}$, with no intercept and a $10^{-8}I_d$ numerical ridge. Initial fitted weights are uniform. Initial centers are sampled from the training inputs without replacement for the heatmap and independently from the true input marginal for the other experiments.

\begin{table}[t]
\centering
\small
\setlength{\tabcolsep}{4pt}

\caption{Settings of the experiments. The dimension is $d=10$, the sample size $N=10{,}000$, and the iteration cap $T=10{,}000$ throughout.}
\begin{tabularx}{\textwidth}{@{}l >{\raggedright\arraybackslash}X
                                  >{\raggedright\arraybackslash}X
                                  >{\raggedright\arraybackslash}X@{}}
\hline
 & Figure~\ref{fig:heatmap}
 & Figure~\ref{fig:warm-start-ablation}
 & Figures~\ref{fig:balance-eff}, \ref{fig:pruning-kl} \\
\hline
true experts $m$
 & $2,\dots,50$
 & $6$
 & $6$ \\
fitted experts $n$
 & $2,4,\dots,100$
 & $4,8,\dots,24$
 & $16$ \\
separation $\Delta$
 & minimum $\geq 6.4$ (scale $8$)
 & $2$, $12$ (exact)
 & $8$ (exact) \\
expert scale $s_\theta$
 & $0.5$
 & $1$
 & $0.5$ \\
mixing weights $\bpi^*$
 & $\operatorname{Dirichlet}(3,\ldots,3)$
 & uniform
 & uniform; skewed \\
expert initialization
 & local least squares
 & local least squares; $\mathcal N(0,I_d)$
 & local least squares \\
window radius $R$
 & $2.5\sqrt{d+\log(2n)}$
 & $2.5\sqrt{d+\log(2n)}$
 & $2.5\sqrt{d+\log(2n)}$ \\
step size $\eta$
 & fixed $0.02$
 & initial $0.02$; backtracking
 & initial $0.02$; backtracking \\
balancing strength $\lambda$
 & --
 & --
 & $0,0.1,0.2,0.4,0.6,0.8,1$ \\
trials / seeds
 & $35$ per cell
 & $20$
 & $20$ \\
success criterion
 & final joint KL $<0.02$ (train)
 & last three joint KL $\leq 0.02$ (test)
 & last three joint KL $\leq 0.02$ (test) \\
\hline
\end{tabularx}
\label{tab:hparams}
\end{table}
\emph{Evaluation.} Since $p_*$ is known, joint KL is estimated by $N_{\mathrm{test}}^{-1}\sum_k\log[p_*(\bz_k)/p(\bz_k)]$ and the conditional KL of Figure~\ref{fig:pruning-kl} by $N_{\mathrm{test}}^{-1}\sum_k\log[p_*(y_k\mid\bx_k)/p(y_k\mid\bx_k)]$, over $N_{\mathrm{test}}=20{,}000$ held-out points in the warm-start, balancing, and pruning experiments. The heatmap instead estimates joint KL on its training points. The other experiments evaluate every $200$ updates; success requires held-out joint KL at most $0.02$ at each of the last three available checkpoints. Balancing models use the full budget. Warm-start and pruning models stop early after three consecutive \emph{validation} joint-KL estimates at most $0.02$, using another independent sample of $20{,}000$ points. Pruning retains the $k$ experts with largest fitted $\pi_i$, renormalizes their weights, and evaluates conditional KL without retraining. Figure~\ref{fig:pruning-kl} plots arithmetic means across all $20$ seeds, including unsuccessful fits, without confidence bands. Success-probability intervals in Figures~\ref{fig:warm-start-ablation} and \ref{fig:balance-eff} are pointwise $95\%$ Wilson intervals; effective-count intervals in Figure~\ref{fig:balance-eff} are pointwise $95\%$ Student-$t$ intervals for the seed mean.

\emph{Compute and code.} The warm-start, balancing, and pruning runs used one NVIDIA A40 GPU, float32 training, and batches of at most $32$ independent trials; initialization and evaluation used float64. Across their full saved grids, recorded amortized GPU training times were $0.17$, $2.39$, and $2.05$ hours, respectively, and recorded evaluation times were $0.25$, $1.11$, and $1.27$ hours. These are measured components rather than end-to-end job wall times, which were not retained. The heatmap runtime was not retained.

\end{document}